\documentclass{article}

\PassOptionsToPackage{numbers, compress}{natbib}

\usepackage[final]{neurips_2021}

\usepackage[utf8]{inputenc}
\usepackage[T1]{fontenc}
\usepackage{url}
\usepackage{booktabs}
\usepackage{amsfonts}
\usepackage{nicefrac}
\usepackage[spacing=true,stretch=10,shrink=20]{microtype}
\usepackage{lipsum}
\usepackage[title]{appendix}
\usepackage{xcolor}
\usepackage{subfig}
\usepackage[newfloat,frozencache]{minted}
\usepackage{xcolor}
\usepackage{xspace}
\usepackage{wrapfig}
\usepackage{cprotect}
\usepackage{enumitem}
\usepackage{siunitx}
\usepackage{comment}
\usepackage[most]{tcolorbox}
\usepackage{authblk}

\definecolor{C0}{HTML}{4C72B0}
\definecolor{C1}{HTML}{DD8452}
\definecolor{C2}{HTML}{55A868}
\definecolor{C3}{HTML}{C44E52}
\definecolor{C4}{HTML}{8172B3}
\definecolor{C5}{HTML}{937860}

\definecolor{laplace1}{RGB}{38, 81, 206}
\definecolor{laplace2}{RGB}{232, 220, 87}
\definecolor{laplace3}{RGB}{238, 238, 238}

\newcommand{\ak}[1]{{\color{C0} [AK: #1]}}

\setminted[python]{
    mathescape,
    linenos,
    numbersep=5pt,
    frame=lines,
    framesep=2mm,
    fontsize=\scriptsize
}

\usepackage[compatibility=false]{caption}
\DeclareCaptionFont{textsc}{\bfseries\selectfont}
\usepackage{hyperref}
\definecolor{mydarkblue}{rgb}{0,0.08,0.45}
\hypersetup{ %
    colorlinks=true,
    linkcolor=mydarkblue,
    citecolor=mydarkblue,
    filecolor=mydarkblue,
    urlcolor=mydarkblue,
}

\usepackage{algorithm}
\usepackage{algpseudocode}

\usepackage{pgfplots}
\pgfplotsset{compat=newest}
\usepgfplotslibrary{groupplots}
\usepgfplotslibrary{dateplot}

\pgfplotsset{compat=1.11,
    /pgfplots/ybar legend/.style={
    /pgfplots/legend image code/.code={%
       \draw[##1,/tikz/.cd,yshift=-0.25em]
        (0cm,0cm) rectangle (3pt,0.8em);},
   },
}

\usepackage{tikz}

\usetikzlibrary{external}
\usetikzlibrary{calc}
\usetikzlibrary{positioning}
\usetikzlibrary{angles,quotes}
\usetikzlibrary{backgrounds}
\usetikzlibrary{fit}
\usetikzlibrary{arrows}
\usetikzlibrary{arrows.meta}
\usetikzlibrary{shapes.symbols}
\usetikzlibrary{shadings}
\usetikzlibrary{shapes}
\usetikzlibrary{fadings}
\usetikzlibrary{bayesnet}
\usetikzlibrary{matrix}
\usetikzlibrary{plotmarks}
\usetikzlibrary{intersections}
\usetikzlibrary{pgfplots.fillbetween}
\pgfplotsset{compat=1.14}
\newcommand*\circled[2]{\tikz[baseline=(char.base)]{\node[shape=circle,draw, #1] (char) {#2};}}

\newcommand*{\mypath}{.}

\usepackage{ifthen}
\usetikzlibrary{shapes,arrows,positioning,calc,chains,scopes}

\definecolor{snowymint}{HTML}{E3F8D1}
\definecolor{wepeep}{HTML}{FAD2D2}
\definecolor{portafino}{HTML}{F5EE9D}
\definecolor{plum}{HTML}{DCACEF}
\definecolor{sail}{HTML}{A3CEEE}
\definecolor{highland}{HTML}{6D885A}

\tikzstyle{signal}=[arrows={},draw=black,line width=0.75pt,rounded corners=4pt]
\tikzstyle{signalstochastic}=[arrows={},draw=laplace1,line width=0.75pt,rounded corners=4pt]

\tikzstyle{block}=[draw=black,line width=1.0pt]
\tikzstyle{cell}=[style=block,draw=highland,rounded corners]
\tikzstyle{celllayer}=[style=block,draw,inner sep=1pt,outer sep=0,
    minimum width=28pt, minimum height=14pt]
\tikzstyle{pointwise}=[style=block,ellipse,inner sep=1pt,outer sep=0, minimum size=12pt]

\tikzstyle{netnode}=[circle, inner sep=0pt, text width=8pt, align=center, line width=1.0pt]
\tikzstyle{inputnode}=[netnode,draw=black,fill=laplace3]
\tikzstyle{hiddennode}=[netnode,draw=black,fill=laplace3]
\tikzstyle{outputnode}=[netnode,draw=black,fill=laplace3]

\def\layerwidth{90pt}
\def\layerheight{14pt}

\tikzstyle{layer}=[style=block, draw, fill=black!20!white,
    inner sep=1pt,outer sep=0, font=\footnotesize,
    text centered,
    minimum width=\layerwidth, minimum height=\layerheight]

\tikzstyle{fc}=[style=layer]
\tikzstyle{conv}=[style=layer, fill=green!30!white]
\tikzstyle{activation}=[style=layer, fill=orange!30!white]
\tikzstyle{pool}=[style=layer, fill=red!30!white]
\tikzstyle{bn}=[style=layer, fill=cyan!30!white]
\tikzstyle{recurrent}=[style=layer, fill=purple!30!white]
\tikzstyle{softmax}=[style=layer, fill=yellow!30!white]
\tikzstyle{point}=[]
\tikzstyle{branch}=[coordinate]

\def\vlayerwidth{30pt}
\def\vlayerheight{3pt}
\def\vblockheight{28pt}

\tikzstyle{vlayer}=[minimum width=\vlayerwidth, minimum height=\vlayerheight]
\tikzstyle{vblock}=[minimum width=\vlayerwidth, minimum height=\vblockheight, text width=1cm, align=center]

\colorlet{fn}{gray!90!green!30!white}
\colorlet{tp}{green!40!white}
\colorlet{fp}{red!40!white}
\colorlet{tn}{gray!90!red!20!white}

\usepackage{amsmath,amsfonts,amssymb,bm}

\def\1{\bm{1}}

\DeclareMathAlphabet{\mathsfit}{\encodingdefault}{\sfdefault}{m}{sl}
\SetMathAlphabet{\mathsfit}{bold}{\encodingdefault}{\sfdefault}{bx}{n}

\newcommand{\E}{\mathbb{E}}

\newcommand{\R}{\mathbb{R}}

\newcommand{\softmax}{\mathrm{softmax}}

\DeclareMathOperator*{\argmax}{arg\,max}
\DeclareMathOperator*{\argmin}{arg\,min}

\newcommand{\MN}{\mathcal{MN}}
\newcommand{\N}{\mathcal{N}}
\newcommand{\diag}[1]{\mathrm{diag}(#1)}

\renewcommand{\R}{\mathbb{R}}
\renewcommand{\E}{\mathop{\mathbb{E}}}
\newcommand{\D}{\mathcal{D}}

\newcommand{\inv}{{-1}}

\renewcommand{\L}{\mathcal{L}}

\newcommand{\defword}[1]{\textbf{\emph{#1}}}
\newcommand{\map}{\text{MAP}}

\graphicspath{{\mypath/figs/}}

\usepackage[capitalise, nameinlink]{cleveref}
\Crefname{appsec}{Appendix}{Appendices}
\Crefname{lstlisting}{Listing}{Listings}

\newcommand{\libname}{\textcolor{blue}{\texttt{laplace}}\xspace}
\renewcommand{\mid}{\ensuremath{\,|\,}}

\title{Laplace Redux -- Effortless Bayesian Deep Learning}

\author[c,m]{Erik Daxberger\thanks{Equal contributors; author ordering sampled uniformly at random.
Correspondence to: \texttt{ead54@cam.ac.uk}, \texttt{agustinus.kristiadi@uni-tuebingen.de}, \texttt{alexander.immer@inf.ethz.ch}, \texttt{runa.eschenhagen@student.uni-tuebingen.de}.}$\;^{,}$}
\author[t]{Agustinus Kristiadi$^{*,}$}
\author[e,p]{Alexander Immer$^{*,}$}
\author[t]{Runa Eschenhagen$^{*,}$}
\author[d]{Matthias Bauer}
\author[t,m]{Philipp Hennig}
\affil[c]{University of Cambridge}
\affil[m]{MPI for Intelligent Systems, T\"{u}bingen}
\affil[t]{University of T\"{u}bingen}
\affil[e]{Department of Computer Science, ETH Zurich}
\affil[p]{Max Planck ETH Center for Learning Systems}
\affil[d]{DeepMind, London}

\makeatletter
\renewcommand{\paragraph}{%
  \@startsection{paragraph}{4}%
  {\z@}{.15ex \@plus 0.15ex \@minus .1ex}{-1em}%
  {\normalfont\normalsize\bfseries}%
}
\makeatother

\usepackage[normalem]{ulem}

\begin{document}

    \maketitle

    \begin{abstract}
      Bayesian formulations of deep learning have been shown to have compelling theoretical properties and offer practical functional benefits, such as improved predictive uncertainty quantification and model selection. The Laplace approximation (LA) is a classic, and arguably the simplest family of approximations for the intractable posteriors of deep neural networks. Yet, despite its simplicity, the LA is not as popular as alternatives like variational Bayes or deep ensembles. This may be due to assumptions that the LA is expensive due to the involved Hessian computation, that it is difficult to implement, or that it yields inferior results.
      In this work we show that these are misconceptions: we (i) review the range of variants of the LA including versions with minimal cost overhead; (ii) introduce \libname{}, an easy-to-use software library for PyTorch offering user-friendly access to all major flavors of the LA; and (iii) demonstrate through extensive experiments that the LA is competitive with more popular alternatives in terms of performance, while excelling in terms of computational cost.
      We hope that this work will serve as a catalyst to a wider adoption of the LA in practical deep learning, including in domains where Bayesian approaches are not typically considered at the moment.
    \end{abstract}
    
    \begin{center}
        \textbf{\libname{} library: \url{https://github.com/AlexImmer/Laplace}}\\
        \textbf{Experiments: \url{https://github.com/runame/laplace-redux}}\\
    \end{center}

    \section{Introduction}
    \label{sec:intro}
    
Despite their successes, modern neural networks (NNs) still suffer from several shortcomings that limit their applicability in some settings.
These include (i) poor calibration and overconfidence, especially when the data distribution shifts between training and testing \citep{guo17calibration}, (ii) catastrophic forgetting of previously learned tasks when continuously trained on new tasks \citep{kirkpatrick2017overcoming}, and (iii) the difficulty of selecting suitable NN architectures and hyperparameters \citep{hutter2019automated}.
Bayesian modeling \citep{barber2012bayesian,ghahramani2015probabilistic} provides a principled and unified approach to tackle these issues by (i) equipping models with robust uncertainty estimates \citep{gal2016dropout}, (ii) enabling models to learn continually by capturing past information \citep{nguyen2018variational}, and (iii) allowing for automated model selection by optimally trading off data fit and model complexity \citep{mackay1995probable}.

Even though this provides compelling motivation for using \emph{Bayesian neural networks} (BNNs) \citep{gal2016uncertainty}, they have not gained much traction in practice. Common criticisms include that BNNs are difficult to implement, finicky to tune, expensive to train, and hard to scale to modern models and datasets. For instance, popular variational Bayesian methods \citep[][etc.]{hinton1993keeping,graves2011practicalVI,blundell2015weight} require considerable changes to the training procedure and model architecture. Also, their optimization process is slower and typically more unstable unless carefully tuned \citep{osawa2019practical}. %
Other methods, such as deep ensembles \citep{lakshminarayanan2017simple}, Monte Carlo dropout \citep{gal2016dropout}, and SWAG \citep{maddox2019simple} promise to bring uncertainty quantification to standard NNs in simple manners. But these methods either require a significant cost increase compared to a single network, have limited empirical performance, or an unsatisfying Bayesian interpretation.

\begin{figure}[!t]

    \centering
    \input{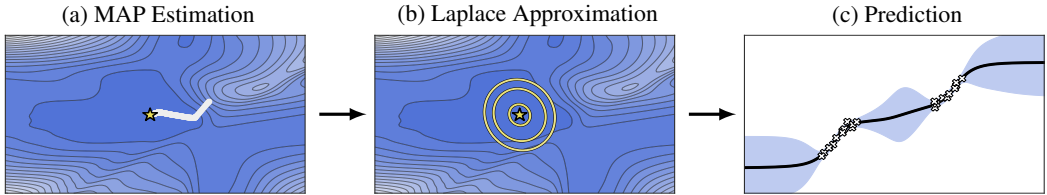}
    \vskip-1.em
    \caption{\textbf{Probabilistic predictions with the Laplace approximation in three steps.} 
    \textbf{(a)} We find a MAP estimate (yellow star) via standard training (background contours = log-posterior landscape on the two-dimensional PCA subspace of the SGD trajectory \citep{izmailov2019subspace}).
    \textbf{(b)} We locally approximate the posterior landscape by fitting a Gaussian centered at the MAP estimate (yellow contours), with covariance matrix equal to the negative inverse Hessian of the loss at the MAP---this is the Laplace approximation (LA).
    \textbf{(c)} We use the LA to make predictions with \emph{predictive uncertainty estimates}---here, the black curve is the predictive mean, and the shading covers the 95\% confidence interval.
    }
    \label{fig:one}
\end{figure}

In this paper, we argue that the Laplace approximation (LA) is a simple and cost-efficient, yet competitive approximation method for inference in Bayesian deep learning. First proposed in this context by \citet{mackay1992bayesian}, the LA dates back to the 18th century \citep{laplace1774memoires}. It
locally approximates the posterior with a Gaussian distribution centered at a local maximum, with covariance matrix corresponding to the local curvature. 
Two key advantages of the LA are that the local maximum is readily available from standard \emph{maximum a posteriori} (MAP) training of NNs, and that curvature estimates can be easily and efficiently obtained thanks to recent advances in second-order optimization, both in terms of more efficient approximations to the Hessian \citep{heskes2000natural,martens2015optimizing,botev2017practical} and easy-to-use software libraries \cite{dangel2020backpack}. Together, they make the LA practical and readily applicable to many already-trained NNs---the LA essentially enables practitioners to turn their high-performing point-estimate NNs into BNNs easily and quickly, without loss of predictive performance. Furthermore, the LA to the marginal likelihood may even be used for Bayesian model selection or NN training \citep{mackay1995probable, immer2021scalable}. \Cref{fig:one} provides an intuition of the LA---we first fit a point estimate of the model and then estimate a Gaussian distribution around that.

Yet, despite recent progress in scaling and improving the LA for deep learning \citep{ritter2018scalable,ritter2018online,khan2019approximate,immer2020improving,daxberger2020expressive,kristiadi2020being,lee2020estimating}, it is %
far less widespread than other methods.
This is likely due to misconceptions, like that the LA is hard to implement due to the Hessian computation, that it must necessarily perform worse than the competitors due to its local nature, or quite simply that 
it is old and too simple.
Here, we show that these are indeed misconceptions.
Moreover, we argue that the LA deserves a wider adoption in both practical and research-oriented deep learning.
To this end, our work makes the following contributions:
\begin{enumerate}
    \item We first survey recent advances and present the key components of scalable and practical Laplace approximations in deep learning (\cref{sec:laplace}).
    \item We then introduce \libname, an easy-to-use PyTorch-based library for ``turning a NN into a BNN'' via the LA (\cref{sec:library}). \libname implements a wide range of different LA variants.
    \item Lastly, using \libname, we show in an extensive empirical study that the LA is competitive to alternative approaches, especially considering how simple and cheap it is (\cref{sec:experiments}).
\end{enumerate}

    \section{The Laplace Approximation in Deep Learning}
    \label{sec:laplace}
    The LA can be used in two different ways to benefit deep learning:
Firstly, we can use the LA to approximate the model's \emph{posterior distribution} (see \cref{eq:laplace_approx} below) to enable \emph{probabilistic predictions} (as also illustrated in \cref{fig:one}).
Secondly, we can use the LA to approximate the \emph{model evidence} (see \cref{eq:laplace_approx_evidence}) to enable \emph{model selection} (e.g.~hyperparameter tuning).

The canonical form of (supervised) deep learning is that of empirical risk minimization. Given, e.g., an i.i.d.\ classification dataset $\D := \{ (x_n \in \R^M, y_n \in \R^C) \}_{n=1}^N$, the weights $\theta\in\R^D$ of an $L$-layer NN $f_\theta: \R^M \to \R^C$ are trained to minimize the (regularized) empirical risk,
which typically decomposes into a sum over empirical loss terms $\ell(x_n,y_n;\theta)$ and a regularizer $r(\theta)$,
\begin{equation}
	\theta_\map = \textstyle\argmin_{\theta \in \R^D} \mathcal{L}(\D; \theta) = \textstyle\argmin_{\theta \in \R^D} \left( r(\theta) + \textstyle\sum_{n=1} ^N \ell(x_n,y_n; \theta) \right).
\end{equation}
From the Bayesian viewpoint, these terms can be identified with i.i.d.~log- \defword{likelihoods} and a log-\defword{prior}, respectively and, thus, $\theta_\map$ is indeed a \defword{maximum a-posteriori (MAP)} estimate:
\begin{equation}
	\ell(x_n,y_n;\theta) = -\log p(y_n\mid f_\theta(x_n)) \qquad\text{and}\qquad r(\theta) = -\log p(\theta)
\end{equation}
For example, the widely used weight regularizer $r(\theta)= \frac{1}{2} \gamma^{-2}\|\theta\|^2$ (a.k.a.\ weight decay) corresponds to a centered Gaussian prior $p(\theta)=\N(\theta;0,\gamma^2 I)$, and the cross-entropy loss
amounts to a categorical likelihood. Hence, the exponential of the negative training loss $\exp(-\mathcal{L}(\D;\theta))$ amounts to an \defword{unnormalized posterior}. By normalizing it, we obtain
\begin{equation}
p(\theta \mid \D) = \tfrac{1}{Z} \,p(\D \mid \theta) \, p(\theta) = \tfrac{1}{Z}\exp(-\mathcal{L}(\D;\theta)), \qquad Z:= \textstyle\int p(\D \mid \theta) \, p(\theta) \,d\theta
\end{equation}
with an intractable \defword{normalizing constant} $Z$. \defword{Laplace approximations} \citep{laplace1774memoires} use a second-order expansion of $\mathcal{L}$ around $\theta_\map$ to construct a Gaussian approximation to $p(\theta \mid \D)$. I.e. we consider:
\begin{equation}
    \mathcal{L}(\D; \theta) \approx \mathcal{L}(\D; \theta_\map) + \tfrac{1}{2} (\theta - \theta_\map)^\intercal \left( \nabla^2 _\theta \mathcal{L}(\D; \theta) \vert_{\theta_\map} \right)(\theta - \theta_\map) ,
\end{equation}
where the first-order term vanishes at $\theta_\map$.
Then we can identify the Laplace approximation as
\tikzexternaldisable
\begin{tcolorbox}[enhanced,colback=white,%
    colframe=C0!75!black, attach boxed title to top right={yshift=-\tcboxedtitleheight/2, xshift=-.75cm}, title=Laplace posterior approximation, coltitle=C0!75!black, boxed title style={size=small,colback=white,opacityback=1, opacityframe=0}, size=title, enlarge top initially by=-\tcboxedtitleheight/2, left=-5pt, after skip=1.5ex plus 0.5ex]
\begin{equation}
	p(\theta \mid \D) \approx \N(\theta; \theta_\map, \varSigma) \qquad\text{with}\qquad \varSigma := \left( \nabla^2_\theta \mathcal{L}(\D;\theta) \vert_{\theta_\map} \right)^\inv.
	\label{eq:laplace_approx}
\end{equation}
\end{tcolorbox}
\tikzexternalenable
The normalizing constant $Z$ (which is typically referred to as the \emph{marginal likelihood} or \emph{evidence}) is useful for model selection and can also be approximated as
\tikzexternaldisable
\begin{tcolorbox}[enhanced,colback=white,%
    colframe=C0!75!black, attach boxed title to top right={yshift=-\tcboxedtitleheight/2, xshift=-.75cm}, title=Laplace approximation of the evidence, coltitle=C0!75!black, boxed title style={size=small,colback=white,opacityback=1, opacityframe=0}, size=title, enlarge top initially by=-\tcboxedtitleheight/2, left=-5pt, after skip=1.5ex plus 0.5ex]
\begin{equation}
	Z \approx \exp(-\mathcal{L}(\D;\theta_\map)) \, (2\pi)^{D/2} \, (\det \varSigma)^{1/2} .
	\label{eq:laplace_approx_evidence}
\end{equation}
\end{tcolorbox}
See \Cref{app:sec:add_details} for more details.
Thus, to obtain the approximate posterior, we first need to find the argmax $\theta_\map$ of the log-posterior function, i.e.~do ``standard'' deep learning with regularized empirical risk minimization. The only \emph{additional} step is to compute the inverse of the Hessian matrix at $\theta_\map$ (see \Cref{fig:one}(b)). The LA can therefore be constructed \emph{post-hoc} to a pre-trained network, even one downloaded off-the-shelf. As we discuss below, the Hessian computation can be offloaded to recently advanced automatic differentiation libraries \citep{dangel2020backpack}.
LAs are widely used to approximate the posterior distribution in logistic regression \citep{spiegelhalter1990sequential}, Gaussian process classification \citep{williams1998bayesian,rasmussen2003gaussian}, and also for Bayesian neural networks (BNNs), both shallow \citep{mackay1992evidence} and deep \citep{ritter2018scalable}. The latter is the focus of this work.

\begin{figure*}[t!]
    \centering

    \input{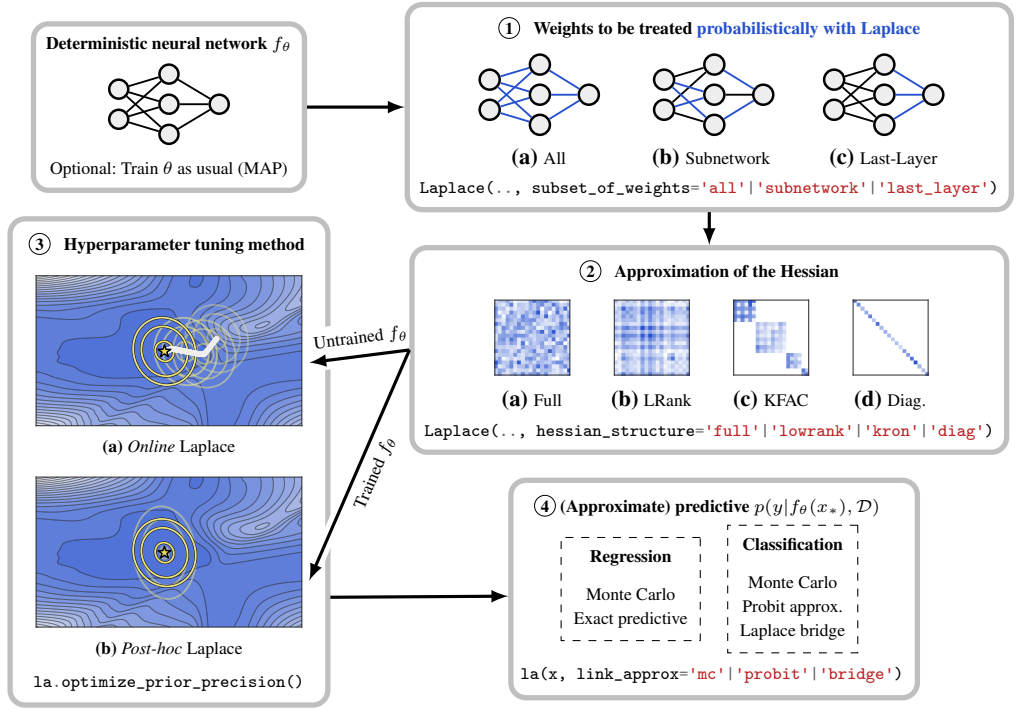}

    \caption[]
    {Four key components
    to scale and apply the LA to a neural network $f_\theta$ (with randomly-initialized or pre-trained weights $\theta$), with corresponding \libname{} code.
    \textbf{\protect\circled{inner sep=1pt}{\small 1}} We first choose which part of the model we want to perform inference over with the LA.
    \textbf{\protect\circled{inner sep=1pt}{\small 2}} We then select how to to approximate the Hessian. 
    \textbf{\protect\circled{inner sep=1pt}{\small 3}} We can then perform model selection using the evidence:
    \textbf{(a)} If we started with an untrained model $f_\theta$, we can jointly train the model and use the evidence to tune hyperparameters \emph{online}.
    \textbf{(b)} If we started with a pre-trained model, we can use the evidence to tune the hyperparameters \emph{post-hoc}.
    Here, shades represent the loss landscape, while contours represent LA log-posteriors---faded contours represent intermediate iterates during hyperparameter tuning to obtain the final log-posterior (thick yellow contours).
    \textbf{\protect\circled{inner sep=1pt}{\small 4}} Finally, to make predictions for a new input $x_*$, we have several options for computing/approximating the predictive distribution $p(y|f_\theta(x_*), \mathcal{D})$.}
    \label{fig:flowchart}
\end{figure*}

Generally, any prior with twice differentiable log-density can be used. Due to the popularity of the weight decay regularizer, we assume that the prior is a zero-mean Gaussian $p(\theta) = \N(\theta; 0, \gamma^2 I)$ unless stated otherwise.%
\footnote{One can also consider a per-layer or even per-parameter weight decay, which corresponds to a more general, but still comparably simple Gaussian prior. In particular, the Hessian of this prior is still diagonal and constant.} 
The Hessian $\nabla^2_\theta \mathcal{L}(\D;\theta) \vert_{\theta_\map}$ then depends both on the (simple) log-prior / regularizer and the (complicated) log-likelihood / empirical risk:
\begin{equation} \label{eq:hessian}
    \nabla^2_\theta \mathcal{L}(\D;\theta) \vert_{\theta_\map} = -\gamma^{-2} I - \textstyle\sum_{n=1}^N \nabla^2_\theta \log p(y_n \mid f_\theta(x_n)) \vert_{\theta_\map}.  %
\end{equation}
A naive implementation of the Hessian is infeasible because the second term in \cref{eq:hessian} scales quadratically with the number of network parameters, which can be in the millions or even billions \citep{he2016deep,shoeybi2019megatron}.
In recent years, several works have addressed scalability, as well as other factors that affect approximation quality and predictive performance of the LA. In the following, we identify, review, and discuss four key components that allow LAs to scale and perform well on modern deep architectures. %
See \cref{fig:flowchart} for an overview and \cref{app:sec:components} for a more detailed version of the review and discussion.

\subsection*{Four Components of Scalable Laplace Approximations for Deep Neural Networks}
\label{subsec:ingredients}

\subsubsection*{\protect\circled{inner sep=1pt}{\small 1}\hspace{0.5em} Inference over all Weights or Subsets of Weights}

In most cases, it is possible to treat \emph{all} weights probabilistically when using appropriate approximations of the Hessian, as we discuss below in \protect\circled{inner sep=1pt}{\small 2}.
Another simple way to scale the LA to large NNs (without Hessian approximations) is the \defword{subnetwork LA} \citep{daxberger2020expressive}, which only treats a \emph{subset} of the model parameters probabilistically with the LA and leaves the remaining parameters at their MAP-estimated values.
An important special case of this applies the LA to only the \emph{last linear layer} of an $L$-layer NN, while fixing the feature extractor defined by the first $L-1$ layers at its MAP estimate~\citep{snoek2015scalable,kristiadi2020being}.
This \defword{last-layer LA} is cost-effective yet compelling both theoretically and in practice~\citep{kristiadi2020being}.

\subsubsection*{\protect\circled{inner sep=1pt}{\small 2}\hspace{0.5em} Hessian Approximations and Their Factorizations}

One advance in second-order optimization that the LA can benefit from are positive semi-definite approximations to the (potentially indefinite) Hessian of the log-likelihoods of NNs in the second term of \cref{eq:hessian} \citep{martens2014new}. The \defword{Fisher information matrix } \citep{amari1998natural}, abbreviated as \emph{the Fisher} and defined by
\begin{equation} \label{eq:FIM}
    F := \textstyle\sum_{n=1}^N \E_{\widehat{y} \sim p(y \mid f_\theta(x_n))} \left[ (\nabla_\theta \log p(\widehat{y} \mid f_\theta(x_n)) \large\vert_{\theta_\map}) (\nabla_\theta \log p(\widehat{y} \mid f_\theta(x_n)) \large\vert_{\theta_\map})^\intercal \right] ,
\end{equation}
is one such choice.\footnote{If, instead of taking expectation in \eqref{eq:FIM}, we use the training label $y_n$, we call the matrix the \defword{empirical Fisher}, which is distinct from the Fisher \citep{martens2014new,kunstner2019limitations}.} One can also use the \defword{generalized Gauss-Newton matrix (GGN)} matrix \citep{schraudolph2002fast}
\begin{equation}
    G := \textstyle\sum_{n=1}^N J(x_n) \left( \nabla^2_{f} \log p(y_n \mid f) \Large\vert_{f=f_{\theta_\map}(x_n)} \right) J(x_n)^\intercal ,
\end{equation}
where $J(x_n) := \nabla_\theta f_\theta(x_n) \vert_{\theta_\map}$ is the NN's Jacobian matrix. As the Fisher and GGN are equivalent for common log-likelihoods~\citep{martens2014new}, we will henceforth refer to them interchangeably. In deep LAs, they have emerged as the default choice \citep[etc.]{ritter2018scalable,ritter2018online,kristiadi2020being,lee2020estimating,daxberger2020expressive,immer2020improving}.

As $F$ and $G$ are still quadratically large, we typically need further factorization assumptions. The most lightweight is a \defword{diagonal factorization} which ignores off-diagonal elements \citep{lecun1990optimal,denker1990transforming}.
More expressive alternatives are block-diagonal factorizations such as
\defword{Kronecker-factored approximate curvature (KFAC)} \citep{heskes2000natural,martens2015optimizing,botev2017practical}, which factorizes each within-layer Fisher\footnote{The elements $F$ or $G$ corresponding to the weight $W_l \subseteq \theta$ of the $l$-th layer of the network.} as a Kronecker product of two smaller matrices. KFAC has been successfully applied to the LA \citep{ritter2018scalable,ritter2018online} and can be improved by low-rank approximations of the KFAC factors \citep{lee2020estimating} by leveraging their eigendecompositions \citep{george2018fast}.
Finally, recent work has studied/enabled \defword{low-rank approximations} of the Hessian/Fisher \citep{madras2019detecting,maddox2020rethinking,sharma2021sketching}.

\subsubsection*{\protect\circled{inner sep=1pt}{\small 3}\hspace{0.5em} Hyperparameter Tuning}

As with all approximate inference methods, the performance of the LA depends on the (hyper)parameters of the prior and likelihood. For instance, it is typically beneficial to tune the prior variance $\gamma^2$ used for inference
\citep{ritter2018scalable,kristiadi2020being,daxberger2020expressive,immer2020improving,immer2021scalable}. Commonly, this is done through \defword{cross-validation}, e.g.~by maximizing the validation log-likelihood \citep{ritter2018scalable,foong2019between} or, additionally, %
using out-of-distribution data \citep{kristiadi2020being,kristiadi2020learnable}.
When using the LA, however, \defword{marginal likelihood maximization} (a.k.a.\ \defword{empirical Bayes} or \defword{the evidence framework} \citep{mackay1992evidence,bernardo2009bayesian}) constitutes a more principled alternative to tune these hyperparameters, and requires no validation data. 
\citet{immer2021scalable} showed that marginal likelihood maximization with LA can work in deep learning and even be performed in an online manner jointly with the MAP estimation.
Note that such approach is not necessarily feasible for other approximate inference methods because most do not provide an estimate of the marginal likelihood.
Other recent approaches for hyperparameter tuning for the LA include Bayesian optimization \citep{humt2020laplaceBO} or the addition of dedicated, trainable hidden units for the sole purpose of uncertainty tuning~\citep{kristiadi2020learnable}.

\subsubsection*{\protect\circled{inner sep=1pt}{\small 4}\hspace{0.5em} Approximate Predictive Distribution}
\label{subsubsec:pred_dist}

To predict using a posterior (approximation) $p(\theta \mid \D)$, we need to compute $p(y \mid f(x_*), \D) = \int p(y \mid f_\theta(x_*)) \, p(\theta \mid \D) \,d\theta$ for any test point $x_* \in \R^n$, which is intractable in general.
The simplest but most general approximation to $p(y \mid x_*, \D)$ is Monte Carlo integration using $S$ samples $(\theta_s)_{s=1}^S$ from $p(\theta \mid \D)$: $p(y \mid f(x_*), \D) \approx S^\inv \sum_{s=1}^S p(y \mid f_{\theta_s}(x_*))$.
However, for LAs with GGN and Fisher Hessian approximations Monte Carlo integration can perform poorly \citep{foong2019between,immer2020improving}. \citet{immer2020improving} attribute this to the inconsistency between Hessian approximation and the predictive and suggest to use a linearized predictive instead, which can also be useful for theoretic analyses~\citep{kristiadi2020being}.
For the last-layer LA, the Hessian coincides with the GGN and the linearized predictive is exact.

The predictive of a \defword{linearized neural network} with a LA approximation to the posterior $p(\theta \mid \D) \approx \N(\theta; \theta_\map, \varSigma)$ results in a Gaussian distribution on neural network outputs $f_* := f(x_*)$ and therefore enables simple approximations or even a closed-form solution.
The distribution on the outputs is given by $p(f_* \mid x_*, \D) \approx \N(f_* ; f_{\theta_\map}(x_*), J(x_*)^\intercal \varSigma J(x_*))$ and is typically significantly lower-dimensional (number of outputs $C$ instead of parameters $D$).
It can also be inferred entirely in function space as a Gaussian process~\citep{khan2019approximate, immer2020improving}.
Given the distribution on outputs $f_*$, the predictive distribution can be obtained by integration against the likelihood:
$p(y \mid x_*, \D) = \int p(y \mid f_*) p(f_* \mid x_*, \D) \, d\theta$.
In the case of regression with a Gaussian likelihood with variance $\sigma^2$, the solution can even be obtained analytically: $p(y \mid x_*, \D) \approx \N(y ; f_{\theta_\map}(x_*), J(x_*)^\intercal \varSigma J(x_*) + \sigma^2 I)$.
For non-Gaussian likelihoods, e.g. in classification, a further approximation is needed.
Again, the simplest approximation to this is \defword{Monte Carlo integration}.
In the binary case, we can employ the \defword{probit approximation} \citep{spiegelhalter1990sequential,mackay1992bayesian} which approximates the logistic function with the probit function.
In the multi-class case, we can use its generalization, the \defword{extended probit approximation} \citep{gibbs1997bayesian}. Finally, first proposed for non-BNN applications \citep{mackay1998choice,hennig2012kernel}, the \defword{Laplace bridge} approximates the softmax-Gaussian integral via a Dirichlet distribution \citep{hobbhahn2020fast}.
The key advantage is that it yields a \emph{distribution} of the integral solutions.

    \section{\libname: A Toolkit for Deep Laplace Approximations}
    \label{sec:library}
    
Implementing the LA is non-trivial, as it requires efficient computation and storage of the Hessian.
While this is not fundamentally difficult, there exists no complete, easy-to-use, and standardized implementation of various LA flavors---%
instead, it is common for deep learning researchers to repeatedly re-implement the LA and Hessian computation with varying efficiency \citep[][etc.]{swag_github,llla_github,dlrrm_github}.
An efficient implementation typically requires hundreds of lines of code, making it hard to quickly prototype with the LA. %
To address this, we introduce \libname{}: a simple, easy-to-use, extensible library for scalable LAs of deep NNs in PyTorch \citep{paszke2019pytorch}.
\libname{} enables \emph{all} sensible combinations of the four components discussed in \cref{subsec:ingredients}---see \cref{fig:flowchart} for details.
\Cref{lst:post_hoc,lst:online} show code examples.

\begin{listing}[t]
    \begin{minipage}[t]{0.47\linewidth}
    \centering

    \begin{minted}[gobble=8]{python}
        from laplace import Laplace

        # Load pre-trained model
        model = load_map_model()

        # Define and fit LA variant with custom settings
        la = Laplace(model, 'classification',
                     subset_of_weights='all', 
                     hessian_structure='diag')
        la.fit(train_loader)
        la.optimize_prior_precision(method='CV',
                                    val_loader=val_loader)

        # Make prediction with custom predictive approx.
        pred = la(x, pred_type='glm', link_approx='probit')
    \end{minted}
    
    \vspace{-0.55em}
    
    \captionof{listing}{Fit diagonal LA over all weights of a pre-trained classification model, do \emph{post-hoc} tuning of the prior precision hyperparameter using cross-validation, and make a prediction for input $x$ with the probit approximation.}
    \label{lst:post_hoc}
    \end{minipage}
    \hfill
    \begin{minipage}[t]{0.47\linewidth}
    \centering

    \begin{minted}[gobble=8]{python}
        from laplace import Laplace

        # Load un- or pre-trained model
        model = load_map_model()  

        # Fit default, recommended LA variant:
        # Last-layer KFAC LA
        la = Laplace(model, 'regression')
        la.fit(train_loader)

        # Differentiate marginal likelihood w.r.t.
        # prior precision and observation noise 
        ml = la.marglik(prior_precision=prior_prec, 
                        sigma_noise=obs_noise)
        ml.backward()
    \end{minted}
    
    \vspace{-0.55em}
    
    \captionof{listing}{Fit KFAC LA over the last layer of a pre- or un-trained regression model and differentiate its marginal likelihood w.r.t. some hyperparameters for \emph{post-hoc} hyperparameter tuning or online empirical Bayes (see \citet{immer2021scalable}).}
    \label{lst:online}
    \end{minipage}
\end{listing}

The core of \libname{} consists of efficient implementations of the LA's key quantities: (i) posterior (i.e.\ Hessian computation and storage), (ii) marginal likelihood, and (iii) posterior predictive. For (i), to take advantage of advances in automatic differentiation, we outsource the Hessian computation to state-of-the-art, optimized second-order optimization libraries: BackPACK \citep{dangel2020backpack} and ASDL \citep{asdl}. Moreover, we design \libname{} in a modular manner that makes it easy to add new backends and approximations in the future. %
For (ii), we follow \citet{immer2021scalable} in our implementation of the LA's marginal likelihood---it is thus both efficient and differentiable and allows the user to implement both \emph{online} and \emph{post-hoc} marginal likelihood tuning, cf. \cref{lst:online}. Note that \libname{} also supports standard cross-validation for hyperparameter tuning \citep{ritter2018scalable,kristiadi2020being}, as shown in \cref{lst:post_hoc}. Finally, for (iii), \libname{} supports all approximations to the posterior predictive distribution discussed in \cref{subsubsec:pred_dist}---it thus provides the user with flexibility in making predictions, depending on the computational budget.

\paragraph*{Default behavior} To abstract away from a large number of options available (\Cref{subsec:ingredients}), we provide the following default choices based on our extensive experiments (\cref{sec:experiments}); they should be applicable and perform decently in the majority of use cases: we assume a pre-trained network and treat only the last-layer weights probabilistically (last-layer LA), use the KFAC factorization of the GGN and tune the hyperparameters \emph{post-hoc} using empirical Bayes. To make predictions, we use the closed-form Gaussian predictive distribution for regression and the (extended) probit approximation for classification. Of course, the user can pick custom choices (\cref{lst:post_hoc,lst:online}).

\paragraph*{Limitations} 
Because \libname{} employs external libraries (BackPACK \citep{dangel2020backpack} and ASDL \citep{asdl}) as backends, it inherits the available choices of Hessian factorizations from these libraries. For instance, the LA variant proposed by \citet{lee2020estimating} can currently not be implemented via \libname{}, because neither backend supports eigenvalue-corrected KFAC \citep{george2018fast} (yet).

    \section{Experiments}
    \label{sec:experiments}
    
We benchmark various LAs implemented via \libname. \Cref{subsec:exp:laplaces} addresses the question of ``which are the best design choices for the LA'', in light of \Cref{fig:flowchart}. \Cref{subsec:exp:uq} shows that the LA is competitive to strong Bayesian baselines in in-distribution, dataset-shift, and out-of-distribution (OOD) settings. We then showcase some applications of the LA in downstream tasks. \Cref{subsec:exp:wilds} demonstrates the applicability of the (last-layer) LA on various data modalities and NN architectures (including transformers \citep{vaswani2017attention})---settings where other Bayesian methods are challenging to implement. \Cref{subsec:exp:online} shows how the LA can be used as an easy-to-use yet strong baseline in continual learning. 
In all results, arrows behind metric names denote if lower ($\downarrow$) or higher ($\uparrow$) values are better.

\subsection{Choosing the Right Laplace Approximation}
\label{subsec:exp:laplaces}

In \cref{sec:laplace} we presented multiple options for each component of the design space of the LA, resulting in a large number of possible combinations, all of which are supported by \libname{}. Here, we try to reduce this complexity and make suggestions for sensible default choices that cover common application scenarios. To this end, we performed a comprehensive comparison between most variants; we measured in- and out-of-distribution performance on standard image classification benchmarks (MNIST, FashionMNIST, CIFAR-10) but also considered the computational complexity of each variant. We provide details of the comparison and a list of the considered variants in \cref{app:sec:la_comparison} and summarize the main arguments and take-aways in the following.

\begin{table}
	\begin{minipage}{0.50\linewidth}
		\centering
        \input{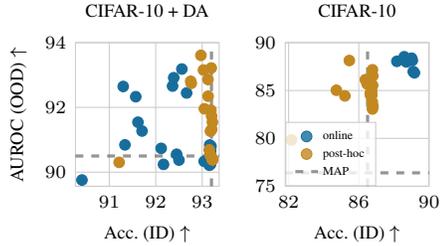}
        
        \captionof{figure}{In- vs. out-of-distribution (ID and OOD, resp.) performance on CIFAR-10 of different LA configurations (dots), each being a combination of settings for 1) subset-of-weights, 2) covariance structure, 3) hyperparameter tuning, and 4) predictive approximation (see \cref{app:sec:la_comparison} for details). 
        ``DA'' stands for ``data augmentation''.
        Post-hoc performs better with DA and a strong pre-trained network, while online performs better without DA where optimal hyperparameters are unknown.}
        \label{fig:comparison}
	\end{minipage}\hfill
	\begin{minipage}{0.47\linewidth}
        \caption{OOD detection performance averaged over all test sets (see \cref{app:sec:uq_details} for details). Confidence is defined as the max.\ of the predictive probability vector \citep{hendrycks17baseline} (e.g. $\text{Confidence}([0.7, 0.2, 0.1]) = 0.7$). LA and especially LA* reduce the overconfidence of MAP and achieve better results than the VB, CSGHMC (\texttt{HMC}), and SWAG (\texttt{SWG}) baselines.}
        \label{tab:ood}
    
        \centering
        \scriptsize
        \renewcommand{\tabcolsep}{4pt}
        \vspace{0.7em}
    
        \begin{tabular}{lrrrrrrr}
            \toprule
    
            & \multicolumn{2}{c}{\bf Confidence $\downarrow$} & \multicolumn{2}{c}{\bf AUROC $\uparrow$} \\
    
            \cmidrule(r){2-3} \cmidrule(r){4-5}
    
            \textbf{Methods} & {\bf MNIST} & {\bf CIFAR-10} & {\bf MNIST} & {\bf CIFAR-10} \\
    
            \midrule
    
            MAP & 75.0$\pm$0.4 & 76.1$\pm$1.2 & 96.5$\pm$0.1 & 92.1$\pm$0.5 \\
            DE & 65.7$\pm$0.3 & 65.4$\pm$0.4 & 97.5$\pm$0.0 & 94.0$\pm$0.1 \\
            VB & 73.2$\pm$0.8 & 58.8$\pm$0.7 & 95.8$\pm$0.2 & 88.7$\pm$0.3 \\
            HMC & 69.2$\pm$1.7 & 69.4$\pm$0.6 & 96.1$\pm$0.2 & 90.6$\pm$0.2 \\
            SWG & 75.8$\pm$0.3 & 68.1$\pm$2.3 & 96.5$\pm$0.1 & 91.3$\pm$0.8 \\
            \midrule
            LA & 67.5$\pm$0.4 & 69.0$\pm$1.3 & 96.2$\pm$0.2 & 92.2$\pm$0.5 \\
            LA* & 56.1$\pm$0.5 & 55.7$\pm$1.2 & 96.4$\pm$0.2 & 92.4$\pm$0.5 \\
    
            \bottomrule
        \end{tabular}
	\end{minipage}
\end{table}

\paragraph{Hyperparameter tuning and parameter inference.}
We can apply the LA purely \emph{post-hoc} (only tune hyperparameters of a pre-trained network) or online (tune hyperparameters and train the network jointly, as e.g. suggested by \citet{immer2021scalable}).
We find that the online LA only works reliably when it is applied to all weights of the network.
In contrast, applying the LA \emph{post-hoc} only on the last layer instead of all weights typically yields better performance due to less underfitting, and is significantly cheaper.
For problems where a pre-trained network or optimal hyperparameters are available, e.g. for well-studied data sets, we, therefore, suggest using the \emph{post-hoc} variant on the last layer.
This LA has the benefit that it has minimal overhead over a standard neural network forward pass~(cf. \cref{fig:costs}) while performing on par or better than state-of-the-art approaches~(cf.~\cref{fig:dataset_shift}).
When hyperparameters are unknown or no validation data is available, we suggest training the neural network online by optimizing the marginal likelihood, following \citet{immer2021scalable} (cf~\cref{subsec:exp:online}).
\Cref{fig:comparison} illustrates this on CIFAR-10:
for CIFAR-10 with data augmentation, strong pre-trained networks and hyperparameters are available and the \emph{post-hoc} methods directly profit from that while the online methods merely reach the same performance.
On the less studied CIFAR-10 without data augmentation, the online method can improve the performance over the \emph{post-hoc} methods.

\paragraph{Covariance approximation and structure.}
Generally, we find that a more expressive covariance approximation improves performance, as would be expected.
However, a full covariance is in most cases intractable for full networks or networks with large last layers.
The KFAC structured covariance provides a good trade-off between expressiveness and speed.
Diagonal approximations perform significantly worse than KFAC and are therefore not suggested.
Independent of the structure, we find that the empirical Fisher (EF) approximations perform better on out-of-distribution detection tasks while GGN approximations tend to perform better on in-distribution metrics.

\paragraph{Predictive distribution.}
Considering in- and out-of-distribution (OOD) performance as well as cost, the probit provides the best approximation to the predictive for the last-layer LA.
MC integration can sometimes be superior for OOD detection but at an increased computational cost.
The Laplace bridge has the same cost as the probit approximation but typically provides inferior results in our experiments.
When using the LA online to optimize hyperparameters, we find that the resulting MAP predictive provides good performance in-distribution, but a probit or MC predictive improves OOD performance.

\paragraph{Overall recommendation.}
Following the experimental evidence, the default in \libname{} is a \emph{post-hoc} KFAC last-layer LA with a GGN approximation to the Hessian.
This default is applicable to all architectures that have a fully-connected last layer and can be easily applied to pre-trained networks.
For problems where trained networks are unavailable or hyperparameters are unknown, the online KFAC LA with a GGN or empirical Fisher provides a good baseline with minimal effort.

\subsection{Predictive Uncertainty Quantification}
\label{subsec:exp:uq}

\begin{figure}[t!]
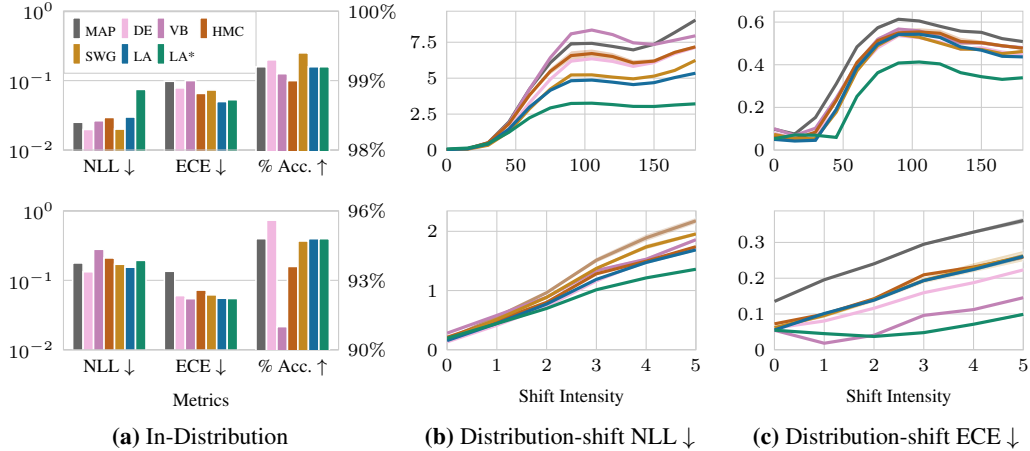

    \def\figfourwidth{0.375\linewidth}
    \def\figfourwidthalt{0.35\linewidth}
    \def\figfourheight{0.15\textheight}

    \subfloat{\begin{tikzpicture}[baseline]

\definecolor{color0}{gray}{0.4}
\definecolor{color1}{rgb}{0.947058823529412,0.723529411764706,0.879411764705882}
\definecolor{color2}{rgb}{0.758823529411765,0.511764705882353,0.711764705882353}
\definecolor{color3}{rgb}{0.730882352941177,0.380882352941177,0.104411764705882}
\definecolor{color4}{rgb}{0.76421568627451,0.531862745098039,0.125980392156863}
\definecolor{color5}{rgb}{0.0906862745098039,0.425980392156863,0.611274509803922}
\definecolor{color6}{rgb}{0.084313725490196,0.543137254901961,0.416666666666667}

\tikzstyle{every node}=[font=\scriptsize]

\begin{axis}[
width=\figfourwidth,
height=\figfourheight,
legend cell align={left},
legend columns=4,
legend style={
  nodes={scale=0.75, transform shape},
  fill opacity=1,
  draw opacity=1,
  text opacity=1,
  at={(0,1)},
  anchor=north west,
  draw=white!80!black
},
axis line style={white!80!black},
tick align=inside,
x grid style={white!80!black},
xmin=-0.5, xmax=2.5,
xtick style={draw=none},
xtick={0,1},
xticklabels={NLL $\downarrow$, ECE $\downarrow$, Acc $\uparrow$},
y grid style={white!80!black},
ymajorgrids,
ytick style={draw=none},
ymode=log,
log basis y={10},
log origin y={infty},
ymin=0.01, ymax=1,
ytick={0.001,0.01,0.1,1,10,100},
]

\draw[draw=white,fill=color0,semithick] (axis cs:-0.4,0.0001) rectangle (axis cs:-0.285714285714286,0.0251519042964093);
\draw[draw=white,fill=color0,semithick] (axis cs:0.6,0.0001) rectangle (axis cs:0.714285714285714,0.0978629158653572);
\addlegendimage{ybar,ybar legend,draw=white,fill=color0,semithick};
\addlegendentry{MAP}

\draw[draw=white,fill=color1,semithick] (axis cs:-0.285714285714286,0.0001) rectangle (axis cs:-0.171428571428571,0.0197959960702807);
\draw[draw=white,fill=color1,semithick] (axis cs:0.714285714285714,0.0001) rectangle (axis cs:0.828571428571429,0.0785662567854392);
\addlegendimage{ybar,ybar legend,draw=white,fill=color1,semithick};
\addlegendentry{DE}

\draw[draw=white,fill=color2,semithick] (axis cs:-0.171428571428571,0.0001) rectangle (axis cs:-0.0571428571428571,0.0264316017746925);
\draw[draw=white,fill=color2,semithick] (axis cs:0.828571428571429,0.0001) rectangle (axis cs:0.942857142857143,0.100195135072324);
\addlegendimage{ybar,ybar legend,draw=white,fill=color2,semithick};
\addlegendentry{VB}

\draw[draw=white,fill=color3,semithick] (axis cs:-0.0571428571428571,0.0001) rectangle (axis cs:0.0571428571428571,0.0294680123448372);
\draw[draw=white,fill=color3,semithick] (axis cs:0.942857142857143,0.0001) rectangle (axis cs:1.05714285714286,0.0654526938906328);
\addlegendimage{ybar,ybar legend,draw=white,fill=color3,semithick};
\addlegendentry{HMC}

\draw[draw=white,fill=color4,semithick] (axis cs:0.0571428571428571,0.0001) rectangle (axis cs:0.171428571428571,0.0200362922906876);
\draw[draw=white,fill=color4,semithick] (axis cs:1.05714285714286,0.0001) rectangle (axis cs:1.17142857142857,0.0733096531439865);
\addlegendimage{ybar,ybar legend,draw=white,fill=color4,semithick};
\addlegendentry{SWG}

\draw[draw=white,fill=color5,semithick] (axis cs:0.171428571428571,0.0001) rectangle (axis cs:0.285714285714286,0.0299043735980988);
\draw[draw=white,fill=color5,semithick] (axis cs:1.17142857142857,0.0001) rectangle (axis cs:1.28571428571429,0.0499586794108253);
\addlegendimage{ybar,ybar legend,draw=white,fill=color5,semithick};
\addlegendentry{LA}

\draw[draw=white,fill=color6,semithick] (axis cs:0.285714285714286,0.0001) rectangle (axis cs:0.4,0.0747756237030029);
\draw[draw=white,fill=color6,semithick] (axis cs:1.28571428571429,0.0001) rectangle (axis cs:1.4,0.0530895099422683);
\addlegendimage{ybar,ybar legend,draw=white,fill=color6,semithick};
\addlegendentry{LA*}

\end{axis}

\begin{axis}[
width=\figfourwidth,
height=\figfourheight,
axis line style={white!80!black},
tick align=inside,
x grid style={white!80!black},
xmin=-0.5, xmax=2.5,
xtick style={draw=none},
xtick={2},
xticklabels={\% Acc. $\uparrow$},
y grid style={white!80!black},
ytick style={draw=none},
axis y line*=right,
ymin=98, ymax=100,
ytick={98, 99, 100},
yticklabels={$98\%$, $99\%$, $100\%$}
]

\draw[draw=white,fill=color0,semithick] (axis cs:1.6,0.0001) rectangle (axis cs:1.71428571428571,99.2);
\draw[draw=white,fill=color1,semithick] (axis cs:1.71428571428571,0.0001) rectangle (axis cs:1.82857142857143,99.3);
\draw[draw=white,fill=color2,semithick] (axis cs:1.82857142857143,0.0001) rectangle (axis cs:1.94285714285714,99.1);
\draw[draw=white,fill=color3,semithick] (axis cs:1.94285714285714,0.0001) rectangle (axis cs:2.05714285714286,99.0);
\draw[draw=white,fill=color4,semithick] (axis cs:2.05714285714286,0.0001) rectangle (axis cs:2.17142857142857,99.4);
\draw[draw=white,fill=color5,semithick] (axis cs:2.17142857142857,0.0001) rectangle (axis cs:2.28571428571429,99.2);
\draw[draw=white,fill=color6,semithick] (axis cs:2.28571428571429,0.0001) rectangle (axis cs:2.4,99.2);
\end{axis}

\end{tikzpicture}}
    \quad\hspace{0.05em}
    \subfloat{\input{\mypath/figs/rmnist_nll}}
    \quad
    \subfloat{\input{\mypath/figs/rmnist_ece}}

    \vspace{-0.75em}
    \setcounter{subfigure}{0}

    \subfloat[In-Distribution]{\begin{tikzpicture}[baseline]

\definecolor{color0}{gray}{0.4}
\definecolor{color1}{rgb}{0.947058823529412,0.723529411764706,0.879411764705882}
\definecolor{color2}{rgb}{0.758823529411765,0.511764705882353,0.711764705882353}
\definecolor{color3}{rgb}{0.730882352941177,0.380882352941177,0.104411764705882}
\definecolor{color4}{rgb}{0.76421568627451,0.531862745098039,0.125980392156863}
\definecolor{color5}{rgb}{0.0906862745098039,0.425980392156863,0.611274509803922}
\definecolor{color6}{rgb}{0.084313725490196,0.543137254901961,0.416666666666667}

\tikzstyle{every node}=[font=\scriptsize]

\begin{axis}[
width=\figfourwidth,
height=\figfourheight,
axis line style={white!80!black},
tick align=inside,
x grid style={white!80!black},
xmin=-0.5, xmax=2.5,
xtick style={draw=none},
xtick={0,1},
xticklabels={NLL $\downarrow$, ECE $\downarrow$, Acc $\uparrow$},
xlabel={Metrics},
y grid style={white!80!black},
ymajorgrids,
ytick style={draw=none},
ymode=log,
log basis y={10},
log origin y={infty},
ymin=0.01, ymax=1,
]

\draw[draw=white,fill=color0,semithick] (axis cs:-0.4,0.0001) rectangle (axis cs:-0.285714285714286,0.17794738945663);
\draw[draw=white,fill=color0,semithick] (axis cs:0.6,0.0001) rectangle (axis cs:0.714285714285714,0.135158108283149);

\draw[draw=white,fill=color1,semithick] (axis cs:-0.285714285714286,0.0001) rectangle (axis cs:-0.171428571428571,0.132984739238024);
\draw[draw=white,fill=color1,semithick] (axis cs:0.714285714285714,0.0001) rectangle (axis cs:0.828571428571429,0.0603124643747098);

\draw[draw=white,fill=color2,semithick] (axis cs:-0.171428571428571,0.0001) rectangle (axis cs:-0.0571428571428571,0.280204570436478);
\draw[draw=white,fill=color2,semithick] (axis cs:0.828571428571429,0.0001) rectangle (axis cs:0.942857142857143,0.0544175366333906);

\draw[draw=white,fill=color3,semithick] (axis cs:-0.0571428571428571,0.0001) rectangle (axis cs:0.0571428571428571,0.210092023468018);
\draw[draw=white,fill=color3,semithick] (axis cs:0.942857142857143,0.0001) rectangle (axis cs:1.05714285714286,0.0723672349516011);

\draw[draw=white,fill=color4,semithick] (axis cs:0.0571428571428571,0.0001) rectangle (axis cs:0.171428571428571,0.1705);
\draw[draw=white,fill=color4,semithick] (axis cs:1.05714285714286,0.0001) rectangle (axis cs:1.17142857142857,0.0618);

\draw[draw=white,fill=color5,semithick] (axis cs:0.171428571428571,0.0001) rectangle (axis cs:0.285714285714286,0.154637783908844);
\draw[draw=white,fill=color5,semithick] (axis cs:1.17142857142857,0.0001) rectangle (axis cs:1.28571428571429,0.055250455570473);

\draw[draw=white,fill=color6,semithick] (axis cs:0.285714285714286,0.0001) rectangle (axis cs:0.4,0.194090641498566);
\draw[draw=white,fill=color6,semithick] (axis cs:1.28571428571429,0.0001) rectangle (axis cs:1.4,0.0547150110239273);

\end{axis}

\begin{axis}[
width=\figfourwidth,
height=\figfourheight,
axis line style={white!80!black},
tick align=inside,
x grid style={white!80!black},
xmin=-0.5, xmax=2.5,
xtick style={draw=none},
xtick={2},
xticklabels={\% Acc. $\uparrow$},
y grid style={white!80!black},
ytick style={draw=none},
axis y line*=right,
ymin=90, ymax=96,
ytick={90, 93, 96},
yticklabels={$90\%$, $93\%$, $96\%$}
]

\draw[draw=white,fill=color0,semithick] (axis cs:1.6,0.0001) rectangle (axis cs:1.71428571428571,94.8);
\draw[draw=white,fill=color1,semithick] (axis cs:1.71428571428571,0.0001) rectangle (axis cs:1.82857142857143,95.6);
\draw[draw=white,fill=color2,semithick] (axis cs:1.82857142857143,0.0001) rectangle (axis cs:1.94285714285714,91.0);
\draw[draw=white,fill=color3,semithick] (axis cs:1.94285714285714,0.0001) rectangle (axis cs:2.05714285714286,93.6);
\draw[draw=white,fill=color4,semithick] (axis cs:2.05714285714286,0.0001) rectangle (axis cs:2.17142857142857,94.7);
\draw[draw=white,fill=color5,semithick] (axis cs:2.17142857142857,0.0001) rectangle (axis cs:2.28571428571429,94.8);
\draw[draw=white,fill=color6,semithick] (axis cs:2.28571428571429,0.0001) rectangle (axis cs:2.4,94.8);

\end{axis}

\end{tikzpicture}}
    \hspace{0.38em}
    \subfloat[Distribution-shift NLL $\downarrow$]{\begin{tikzpicture}[trim axis right, baseline]

\definecolor{color0}{rgb}{0.740686274509804,0.573039215686274,0.431862745098039}
\definecolor{color1}{rgb}{0.947058823529412,0.723529411764706,0.879411764705882}
\definecolor{color2}{rgb}{0.758823529411765,0.511764705882353,0.711764705882353}
\definecolor{color3}{rgb}{0.730882352941177,0.380882352941177,0.104411764705882}
\definecolor{color4}{rgb}{0.76421568627451,0.531862745098039,0.125980392156863}
\definecolor{color5}{rgb}{0.0906862745098039,0.425980392156863,0.611274509803922}
\definecolor{color6}{rgb}{0.084313725490196,0.543137254901961,0.416666666666667}

\tikzstyle{every node}=[font=\scriptsize]

\begin{axis}[
width=\figfourwidthalt,
height=\figfourheight,
axis line style={white!80!black},
tick align=inside,
x grid style={white!80!black},
xmajorgrids,
xlabel={Shift Intensity},
xmin=0, xmax=5,
xtick style={draw=none},
xtick={0,1,2,3,4,5},
y grid style={white!80!black},
ymajorgrids,
ymin=0, ymax=2.34484968957857,
ytick style={draw=none},
ytick={0,1,2,3},
]

\path [draw=white, fill=color0, opacity=0.3, semithick]
(axis cs:0,0.181417808028295)
--(axis cs:0,0.174476970884964)
--(axis cs:1,0.543053364613925)
--(axis cs:2,0.929334805386673)
--(axis cs:3,1.45725114789172)
--(axis cs:4,1.82106721514216)
--(axis cs:5,2.1154572648871)
--(axis cs:5,2.23946517330208)
--(axis cs:5,2.23946517330208)
--(axis cs:4,1.95923429062693)
--(axis cs:3,1.5665839437389)
--(axis cs:2,0.998773602253784)
--(axis cs:1,0.568017336155102)
--(axis cs:0,0.181417808028295)
--cycle;

\path [draw=white, fill=color1, opacity=0.3, semithick]
(axis cs:0,0.134194630703577)
--(axis cs:0,0.13177484777247)
--(axis cs:1,0.408756166914147)
--(axis cs:2,0.721810798042895)
--(axis cs:3,1.14911953999585)
--(axis cs:4,1.48051371102179)
--(axis cs:5,1.70828857999176)
--(axis cs:5,1.73205350096056)
--(axis cs:5,1.73205350096056)
--(axis cs:4,1.51781857626274)
--(axis cs:3,1.17553830886457)
--(axis cs:2,0.741130288003801)
--(axis cs:1,0.417770375920294)
--(axis cs:0,0.134194630703577)
--cycle;

\path [draw=white, fill=color2, opacity=0.3, semithick]
(axis cs:0,0.282804002136723)
--(axis cs:0,0.277605138736232)
--(axis cs:1,0.573760593325563)
--(axis cs:2,0.8704441089387)
--(axis cs:3,1.30412372471744)
--(axis cs:4,1.48876493629982)
--(axis cs:5,1.81550439571145)
--(axis cs:5,1.90110273788052)
--(axis cs:5,1.90110273788052)
--(axis cs:4,1.5687054987664)
--(axis cs:3,1.37187248162176)
--(axis cs:2,0.902589064285431)
--(axis cs:1,0.586608893602423)
--(axis cs:0,0.282804002136723)
--cycle;

\path [draw=white, fill=color3, opacity=0.3, semithick]
(axis cs:0,0.213382495120828)
--(axis cs:0,0.206801551815207)
--(axis cs:1,0.488693644715645)
--(axis cs:2,0.784866651001183)
--(axis cs:3,1.26682647583596)
--(axis cs:4,1.46854650261916)
--(axis cs:5,1.707646664417)
--(axis cs:5,1.77301543070057)
--(axis cs:5,1.77301543070057)
--(axis cs:4,1.52312387944662)
--(axis cs:3,1.29869105760781)
--(axis cs:2,0.816715620818726)
--(axis cs:1,0.498209417676431)
--(axis cs:0,0.213382495120828)
--cycle;

\path [draw=white, fill=color4, opacity=0.3, semithick]
(axis cs:0,0.172826370563775)
--(axis cs:0,0.168173629436225)
--(axis cs:1,0.50362887577762)
--(axis cs:2,0.862368003115265)
--(axis cs:3,1.34067565102856)
--(axis cs:4,1.68396337802265)
--(axis cs:5,1.91392165640142)
--(axis cs:5,1.99807834359858)
--(axis cs:5,1.99807834359858)
--(axis cs:4,1.78863662197735)
--(axis cs:3,1.40768434897144)
--(axis cs:2,0.908591996884735)
--(axis cs:1,0.52133112422238)
--(axis cs:0,0.172826370563775)
--cycle;

\path [draw=white, fill=color5, opacity=0.3, semithick]
(axis cs:0,0.157153144525709)
--(axis cs:0,0.152122423291979)
--(axis cs:1,0.43551819770561)
--(axis cs:2,0.753418540236434)
--(axis cs:3,1.15497305086728)
--(axis cs:4,1.43338070797543)
--(axis cs:5,1.65238718025092)
--(axis cs:5,1.72090899650689)
--(axis cs:5,1.72090899650689)
--(axis cs:4,1.51464760501285)
--(axis cs:3,1.22154253384952)
--(axis cs:2,0.78965200419716)
--(axis cs:1,0.453453750231402)
--(axis cs:0,0.157153144525709)
--cycle;

\path [draw=white, fill=color6, opacity=0.3, semithick]
(axis cs:0,0.195801276928617)
--(axis cs:0,0.192380006068515)
--(axis cs:1,0.437102892750733)
--(axis cs:2,0.687014850898895)
--(axis cs:3,0.996130304126015)
--(axis cs:4,1.1909279742703)
--(axis cs:5,1.34265303010279)
--(axis cs:5,1.37689793059056)
--(axis cs:5,1.37689793059056)
--(axis cs:4,1.23631174496798)
--(axis cs:3,1.03094305646969)
--(axis cs:2,0.708233560355988)
--(axis cs:1,0.447963753367431)
--(axis cs:0,0.195801276928617)
--cycle;

\addplot [very thick, color0]
table {%
0 0.17794738945663
1 0.555535350384514
2 0.964054203820229
3 1.51191754581531
4 1.89015075288455
5 2.17746121909459
};
\addplot [very thick, color1]
table {%
0 0.132984739238024
1 0.41326327141722
2 0.731470543023348
3 1.16232892443021
4 1.49916614364227
5 1.72017104047616
};
\addplot [very thick, color2]
table {%
0 0.280204570436478
1 0.580184743463993
2 0.886516586612066
3 1.3379981031696
4 1.52873521753311
5 1.85830356679598
};
\addplot [very thick, color3]
table {%
0 0.210092023468018
1 0.493451531196038
2 0.800791135909955
3 1.28275876672188
4 1.49583519103289
5 1.74033104755878
};
\addplot [very thick, color4]
table {%
0 0.1705
1 0.51248
2 0.88548
3 1.37418
4 1.7363
5 1.956
};
\addplot [very thick, color5]
table {%
0 0.154637783908844
1 0.444485973968506
2 0.771535272216797
3 1.1882577923584
4 1.47401415649414
5 1.68664808837891
};
\addplot [very thick, color6]
table {%
0 0.194090641498566
1 0.442533323059082
2 0.697624205627441
3 1.01353668029785
4 1.21361985961914
5 1.35977548034668
};
\end{axis}

\end{tikzpicture}}
    \quad
    \subfloat[Distribution-shift ECE $\downarrow$]{\begin{tikzpicture}[baseline]

\definecolor{color0}{gray}{0.4}
\definecolor{color1}{rgb}{0.947058823529412,0.723529411764706,0.879411764705882}
\definecolor{color2}{rgb}{0.758823529411765,0.511764705882353,0.711764705882353}
\definecolor{color3}{rgb}{0.730882352941177,0.380882352941177,0.104411764705882}
\definecolor{color4}{rgb}{0.76421568627451,0.531862745098039,0.125980392156863}
\definecolor{color5}{rgb}{0.0906862745098039,0.425980392156863,0.611274509803922}
\definecolor{color6}{rgb}{0.084313725490196,0.543137254901961,0.416666666666667}

\tikzstyle{every node}=[font=\scriptsize]

\begin{axis}[
width=\figfourwidthalt,
height=\figfourheight,
axis line style={white!80!black},
tick align=inside,
x grid style={white!80!black},
xmajorgrids,
xlabel={Shift Intensity},
xmin=0, xmax=5,
xtick style={draw=none},
xtick={0,1,2,3,4,5},
y grid style={white!80!black},
ymajorgrids,
ytick style={draw=none},
ymin=0, ymax=0.388155754288103,
ytick={0,0.1,0.2,0.3,0.4},
]

\path [draw=white, fill=color0, opacity=0.3, semithick]
(axis cs:0,0.13806564615997)
--(axis cs:0,0.132250570406327)
--(axis cs:1,0.191823037826692)
--(axis cs:2,0.234853175946613)
--(axis cs:3,0.289043157501096)
--(axis cs:4,0.322299195523617)
--(axis cs:5,0.351952765014447)
--(axis cs:5,0.37050923169012)
--(axis cs:5,0.37050923169012)
--(axis cs:4,0.33561701936673)
--(axis cs:3,0.300912594558578)
--(axis cs:2,0.245695203244278)
--(axis cs:1,0.199491854686255)
--(axis cs:0,0.13806564615997)
--cycle;

\path [draw=white, fill=color1, opacity=0.3, semithick]
(axis cs:0,0.0615826821138649)
--(axis cs:0,0.0590422466355546)
--(axis cs:1,0.0792579647545795)
--(axis cs:2,0.114301562914632)
--(axis cs:3,0.157336535293163)
--(axis cs:4,0.184219630030102)
--(axis cs:5,0.220240748614775)
--(axis cs:5,0.226262555664104)
--(axis cs:5,0.226262555664104)
--(axis cs:4,0.190776814459027)
--(axis cs:3,0.161623368274936)
--(axis cs:2,0.117873319194608)
--(axis cs:1,0.08159117898387)
--(axis cs:0,0.0615826821138649)
--cycle;

\path [draw=white, fill=color2, opacity=0.3, semithick]
(axis cs:0,0.0550193327927994)
--(axis cs:0,0.0538157404739818)
--(axis cs:1,0.0175787797304617)
--(axis cs:2,0.0379977632615836)
--(axis cs:3,0.0918442671829485)
--(axis cs:4,0.108271178562712)
--(axis cs:5,0.141403408812152)
--(axis cs:5,0.149652991066443)
--(axis cs:5,0.149652991066443)
--(axis cs:4,0.11639855986863)
--(axis cs:3,0.100389431648411)
--(axis cs:2,0.0430560566994878)
--(axis cs:1,0.0188953410626318)
--(axis cs:0,0.0550193327927994)
--cycle;

\path [draw=white, fill=color3, opacity=0.3, semithick]
(axis cs:0,0.0755852202442326)
--(axis cs:0,0.0691492496589696)
--(axis cs:1,0.0982900263118335)
--(axis cs:2,0.138995432889503)
--(axis cs:3,0.207288691540045)
--(axis cs:4,0.226622516296519)
--(axis cs:5,0.253094285348384)
--(axis cs:5,0.265209675843719)
--(axis cs:5,0.265209675843719)
--(axis cs:4,0.235785649267983)
--(axis cs:3,0.211825258923909)
--(axis cs:2,0.145933186159362)
--(axis cs:1,0.10067012075619)
--(axis cs:0,0.0755852202442326)
--cycle;

\path [draw=white, fill=color4, opacity=0.3, semithick]
(axis cs:0,0.068295382975622)
--(axis cs:0,0.055304617024378)
--(axis cs:1,0.0874746124711355)
--(axis cs:2,0.130852094612269)
--(axis cs:3,0.183651703005934)
--(axis cs:4,0.214652266705452)
--(axis cs:5,0.245525689170727)
--(axis cs:5,0.278354310829273)
--(axis cs:5,0.278354310829273)
--(axis cs:4,0.243987733294548)
--(axis cs:3,0.203908296994066)
--(axis cs:2,0.149707905387731)
--(axis cs:1,0.103925387528865)
--(axis cs:0,0.068295382975622)
--cycle;

\path [draw=white, fill=color5, opacity=0.3, semithick]
(axis cs:0,0.0560819755359691)
--(axis cs:0,0.054418935604977)
--(axis cs:1,0.0984696650629177)
--(axis cs:2,0.13554034146532)
--(axis cs:3,0.186903686746048)
--(axis cs:4,0.21779800140355)
--(axis cs:5,0.251237387007336)
--(axis cs:5,0.270427623434967)
--(axis cs:5,0.270427623434967)
--(axis cs:4,0.23134875769723)
--(axis cs:3,0.199997941356741)
--(axis cs:2,0.14179806452393)
--(axis cs:1,0.104101265016976)
--(axis cs:0,0.0560819755359691)
--cycle;

\path [draw=white, fill=color6, opacity=0.3, semithick]
(axis cs:0,0.0556163009521181)
--(axis cs:0,0.0538137210957366)
--(axis cs:1,0.044038514610405)
--(axis cs:2,0.0360328473056511)
--(axis cs:3,0.0454519947577703)
--(axis cs:4,0.0671233140954189)
--(axis cs:5,0.0924595501406901)
--(axis cs:5,0.105431082309771)
--(axis cs:5,0.105431082309771)
--(axis cs:4,0.0755997294236457)
--(axis cs:3,0.0503468540459313)
--(axis cs:2,0.0383356654130861)
--(axis cs:1,0.0461287339582766)
--(axis cs:0,0.0556163009521181)
--cycle;

\addplot [very thick, color0]
table {%
0 0.135158108283149
1 0.195657446256474
2 0.240274189595445
3 0.294977876029837
4 0.328958107445173
5 0.361230998352284
};
\addplot [very thick, color1]
table {%
0 0.0603124643747098
1 0.0804245718692248
2 0.11608744105462
3 0.159479951784049
4 0.187498222244564
5 0.223251652139439
};
\addplot [very thick, color2]
table {%
0 0.0544175366333906
1 0.0182370603965467
2 0.0405269099805357
3 0.0961168494156798
4 0.112334869215671
5 0.145528199939297
};
\addplot [very thick, color3]
table {%
0 0.0723672349516011
1 0.0994800735340118
2 0.142464309524432
3 0.209556975231977
4 0.231204082782251
5 0.259151980596051
};
\addplot [very thick, color4]
table {%
0 0.0618
1 0.0957
2 0.14028
3 0.19378
4 0.22932
5 0.26194
};
\addplot [very thick, color5]
table {%
0 0.055250455570473
1 0.101285465039947
2 0.138669202994625
3 0.193450814051394
4 0.22457337955039
5 0.260832505221151
};
\addplot [very thick, color6]
table {%
0 0.0547150110239273
1 0.0450836242843408
2 0.0371842563593686
3 0.0478994244018508
4 0.0713615217595323
5 0.0989453162252308
};
\end{axis}

\end{tikzpicture}}

    \caption{Assessing model calibration \textbf{(a)} on in-distribution data and \textbf{(b,c)} under distribution shift, for the MNIST (top row) and CIFAR-10 (bottom row) datasets. 
    For \textbf{(b,c)}, we use the Rotated-MNIST (top) and Corrupted-CIFAR-10 (bottom) benchmarks \citep{hendrycks2018benchmarking,ovadia2019can}.
    In \textbf{(a)}, we report accuracy and, to measure calibration, negative log-likelihood (NLL) and expected calibration error (ECE)---all evaluated on the standard test sets.
    In \textbf{(b)} and \textbf{(c)}, we plot shift intensities against NLL and ECE, respectively.
    For Rotated-MNIST (top), shift intensities denote degrees of rotation of the images, while for Corrupted-CIFAR-10 (bottom), they denote the amount of image distortion (see \cite{hendrycks2018benchmarking,ovadia2019can} for details). 
    \textbf{(a)} On in-distribution data, LA is the best-calibrated method in terms of ECE, while also retaining the accuracy of MAP (unlike VB and CSGHMC).
    \textbf{(b,c)} On corrupted data, all Bayesian methods improve upon MAP significantly. Even though \emph{post-hoc}, all LAs achieve competitive results, even to DE. In particular, LA* achieves the best results, at the expense of slightly worse in-distribution calibration---this trade-off between in- and out-of-distribution performance has been observed previously \citep{lu2020uncertainty}.
    }
    \label{fig:dataset_shift}
\end{figure}

We consider two flavors of LAs: the default flavor of \libname (\textbf{LA}) and the most robust one in terms of distribution shift found in \cref{subsec:exp:laplaces} (\textbf{LA*}---last-layer, with a full empirical Fisher Hessian approximation, and the probit approximation). We compare them with the MAP network (\textbf{MAP}) and various popular and strong Bayesian baselines: Deep Ensemble \citep[\textbf{DE},][]{lakshminarayanan2017simple}, mean-field variational Bayes \citep[\textbf{VB},][]{graves2011practicalVI,blundell2015weight} with the flipout estimator \citep{wen2018flipout}, cyclical stochastic-gradient Hamiltonian Monte Carlo \citep[\textbf{CSGHMC / HMC},][]{zhang2020csgmcmc}, and SWAG \citep[\textbf{SWG},][]{maddox2019simple}. For each baseline, we use the hyperparameters recommended in the original paper---see \cref{app:sec:add_details} for details.
First, \cref{fig:dataset_shift} shows that \textbf{LA} and \textbf{LA*} are, respectively, competitive with and superior to the baselines in trading-off between in-distribution calibration and dataset-shift robustness.
Second, \cref{tab:ood} shows that \textbf{LA} and \textbf{LA*} achieve better results on out-of-distribution (OOD) detection than even \textbf{VB}, \textbf{CSGHMC}, and \textbf{SWG}. 

\begin{wrapfigure}[10]{r}{0.4\textwidth}
    \centering

    \vspace{-0.8em}
    \begin{tikzpicture}

\definecolor{color0}{rgb}{0.0906862745098039,0.425980392156863,0.611274509803922}
\definecolor{color1}{rgb}{0.76421568627451,0.531862745098039,0.125980392156863}
\definecolor{color2}{rgb}{0.084313725490196,0.543137254901961,0.416666666666667}
\definecolor{color3}{rgb}{0.730882352941177,0.380882352941177,0.104411764705882}
\definecolor{color4}{rgb}{0.758823529411765,0.511764705882353,0.711764705882353}
\definecolor{color5}{rgb}{0.740686274509804,0.573039215686274,0.431862745098039}
\definecolor{color6}{rgb}{0.947058823529412,0.723529411764706,0.879411764705882}

\tikzstyle{every node}=[font=\scriptsize]

\begin{axis}[
width=\linewidth,
height=0.15\textheight,
legend style={nodes={scale=0.75, transform shape}, fill opacity=1, draw opacity=1, text opacity=1, draw=white!80!black, anchor=north west, at={(0, 1)}},
legend cell align={left},
axis line style={white!80!black},
tick align=inside,
xmin=-0.5, xmax=5.5,
xtick style={draw=none},
xtick={0,1,2,3,4,5},
xticklabels={MAP,DE,VB,HMC,SWG,LA},
y grid style={white!80!black},
ylabel={Relative Time $\downarrow$},
ymajorgrids,
ymin=0, ymax=13.52941593123,
ytick={0,2,5,10},
ytick style={draw=none}
]
\draw[draw=white,fill=color0,semithick] (axis cs:-0.4,0) rectangle (axis cs:0,1);
\addlegendimage{ybar,ybar legend,draw=white,fill=color0,semithick};
\addlegendentry{Training}

\draw[draw=white,fill=color0,semithick] (axis cs:0.6,0) rectangle (axis cs:1,4.1604198951211);
\draw[draw=white,fill=color0,semithick] (axis cs:1.6,0) rectangle (axis cs:2,2.08843032626727);
\draw[draw=white,fill=color0,semithick] (axis cs:2.6,0) rectangle (axis cs:3,4.28123499725073);
\draw[draw=white,fill=color0,semithick] (axis cs:3.6,0) rectangle (axis cs:4,1.73364000917944);
\draw[draw=white,fill=color0,semithick] (axis cs:4.6,0) rectangle (axis cs:5,1.01500699050791);
\draw[draw=white,fill=color1,semithick] (axis cs:-2.77555756156289e-17,0) rectangle (axis cs:0.4,1);
\addlegendimage{ybar,ybar legend,draw=white,fill=color1,semithick};
\addlegendentry{Prediction}

\draw[draw=white,fill=color1,semithick] (axis cs:1,0) rectangle (axis cs:1.4,2.49909270132756);
\draw[draw=white,fill=color1,semithick] (axis cs:2,0) rectangle (axis cs:2.4,4.41139174490005);
\draw[draw=white,fill=color1,semithick] (axis cs:3,0) rectangle (axis cs:3.4,5.1530602591538);
\draw[draw=white,fill=color1,semithick] (axis cs:4,0) rectangle (axis cs:4.4,12.8851580297428);
\draw[draw=white,fill=color1,semithick] (axis cs:5,0) rectangle (axis cs:5.4,1);
\end{axis}

\end{tikzpicture}

    \caption{Wall-clock time costs relative to MAP. LA introduces negligible overhead over MAP, while all other baselines are significantly more expensive.}
    \label{fig:costs}
\end{wrapfigure}
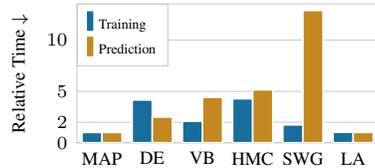

The LA shines even more when we consider its (time \emph{and} memory) cost relative to the other, more complex baselines. 
In \cref{fig:costs} we show the wall-clock times of each method relative to \textbf{MAP}'s for training and prediction. As expected, \textbf{DE}, \textbf{VB}, and \textbf{CSGHMC} are slow to train and in making predictions: they are between two to five times more expensive than MAP. Meanwhile, despite being \emph{post-hoc}, \textbf{SWG} is almost twice as expensive as \textbf{MAP} during training due to the need for sampling and updating its batch normalization statistics. Moreover, with 30 samples, as recommended by its authors \citep{maddox2019simple}, it is very expensive at prediction time---more than ten times more expensive than \textbf{MAP}.
Meanwhile, \textbf{LA} (and \textbf{LA*}) is the cheapest of all methods considered: it only incurs a negligible overhead on top of the costs of \textbf{MAP}.
This is similar for the memory consumption (see \cref{tab:memory} in \cref{app:sec:memory}).
This shows that the LA is significantly more memory- and compute-efficient than all the other methods, adding minimal overhead over MAP inference and prediction.
This makes the LA particularly attractive for practitioners, especially in low-resource environments.
Together with \cref{fig:dataset_shift,tab:ood}, this justifies our default flavor in \libname{}, and importantly, shows that Bayesian deep learning does not have to be expensive.%

\subsection{Realistic Distribution Shift}
\label{subsec:exp:wilds}

\begin{figure}[t!]
    \def\figsixwidth{0.175\linewidth}
    \def\figsixheight{0.15\textheight}

    \hspace{36em}
    \input{\mypath/figs/wilds_laplace}
    
    \vspace{-3mm}

    \hspace{.03\linewidth}
    \subfloat[\label{subfig:camelyon17} \texttt{Camelyon17}]{\hspace{.19\linewidth}}
    \subfloat[\label{subfig:fmow} \texttt{FMoW}]{\hspace{.19\linewidth}}
    \subfloat[\label{subfig:civilcomments} \texttt{CivilComments}]{\hspace{.20\linewidth}}
    \subfloat[\label{subfig:amazon} \texttt{Amazon}]{\hspace{.19\linewidth}}
    \subfloat[\label{subfig:poverty} \texttt{PovertyMap}]{\hspace{.19\linewidth}}

    \caption{Assessing real-world distribution shift robustness on five datasets from the \texttt{WILDS} benchmark \cite{koh2020wilds}, covering different data modalities, model architectures, and output types.
    \texttt{Camelyon17}: Tissue slide image tumor classification across hospitals (DenseNet-121 \cite{huang2017densely}).
    \texttt{FMoW}: Satellite image land use classification across regions/years (DenseNet-121).
    \texttt{CivilCommments}: Online comment toxicity classification across demographics (DistilBERT \cite{sanh2019distilbert}).
    \texttt{Amazon}: Product review sentiment classification across users (DistilBERT).
    \texttt{PovertyMap}: Satellite image asset wealth regression across countries (ResNet-18 \cite{he2016deep}).
    We plot means $\pm$ standard errors of the NLL (top) and ECE (for classification) or regression calibration error \cite{kuleshov2018accurate} (bottom).
    The in-distribution (left panels) and OOD (right panels) dataset splits correspond to different domains (e.g.\ hospitals for \texttt{Camelyon17}).
    LA is much better calibrated than MAP, and competitive with temp.~scaling and DE, especially on the OOD splits.
    }
    \label{fig:wilds}
    \vspace{-0.5em}
\end{figure}

So far, our experiments focused on comparably simple benchmarks, allowing us to comprehensively assess different LA variants and compare to more involved Bayesian methods such as VB, MCMC, and SWAG.
In more realistic settings, however, where we want to improve the uncertainty of complex and costly-to-train models, such as transformers \cite{vaswani2017attention}, these methods would likely be difficult to get to work well and expensive to run.
However, one might often have access to a pre-trained model, allowing for the cheap use of \emph{post-hoc} methods such as the LA.
To demonstrate this, we show how \libname can improve the distribution shift robustness of complex pre-trained models in large-scale settings.
To this end, we use \texttt{WILDS} \cite{koh2020wilds}, a recently proposed benchmark of realistic distribution shifts encompassing a variety of real-world datasets across different data modalities and application domains.
While the \texttt{WILDS} models employ complex (e.g.\ convolutional or transformer) architectures as feature extractors, they all feed into a linear output layer, allowing us to conveniently and cheaply apply the last-layer LA.
As baselines, we consider:
1) the pre-trained MAP models \cite{koh2020wilds}, 2) \emph{post-hoc} temperature scaling of the MAP models (for classification tasks) \cite{guo17calibration}, and 3) deep ensembles \cite{lakshminarayanan2017simple}.\footnote{We simply construct deep ensembles from the various pre-trained models provided by \citet{koh2020wilds}.}
More details on the experimental setup are provided in \cref{sec:wilds_details}.
\cref{fig:wilds} shows the results on five different \texttt{WILDS} datasets (see caption for details).
Overall, Laplace is significantly better calibrated than MAP, and competitive with temperature scaling and ensembles, especially on the OOD splits.

\subsection{Further Applications}
\label{subsec:exp:online}

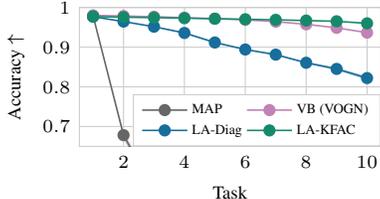
\begin{wrapfigure}[14]{r}{0.4\textwidth}
    \centering
    \vspace{-1.75em}

    \begin{tikzpicture}

\definecolor{color0}{gray}{0.4}
\definecolor{color1}{rgb}{0.758823529411765,0.511764705882353,0.711764705882353}
\definecolor{color2}{rgb}{0.0906862745098039,0.425980392156863,0.611274509803922}
\definecolor{color3}{rgb}{0.084313725490196,0.543137254901961,0.416666666666667}

\tikzstyle{every node}=[font=\scriptsize]

\begin{axis}[
width=\linewidth,
height=0.15\textheight,
axis line style={white!80!black},
legend cell align={left},
legend columns=2, 
legend style={nodes={scale=0.75, transform shape}, fill opacity=1, draw opacity=1, text opacity=1, draw=white!80!black, anchor=south east, at={(1, 0)}},
tick align=inside,
x grid style={white!80!black},
xlabel={Task},
xmajorgrids,
xmin=0.55, xmax=10.45,
xtick style={draw=none},
y grid style={white!80!black},
ylabel={Accuracy $\uparrow$},
ymajorgrids,
ymin=0.65, ymax=1.0,
ytick={0.6, 0.7, 0.8, 0.9, 1},
ytick style={draw=none}
]
\path [draw=white, fill=color0, opacity=0.3]
(axis cs:1,0.978085393178987)
--(axis cs:1,0.979674606821013)
--(axis cs:2,0.69145634289153)
--(axis cs:3,0.549230794500681)
--(axis cs:4,0.425915825472957)
--(axis cs:5,0.378694502737166)
--(axis cs:6,0.342587940729112)
--(axis cs:7,0.301384910517227)
--(axis cs:8,0.265188541511148)
--(axis cs:9,0.268461883138065)
--(axis cs:10,0.247768172029709)
--(axis cs:10,0.230991827970291)
--(axis cs:10,0.230991827970291)
--(axis cs:9,0.25118256130638)
--(axis cs:8,0.258211458488852)
--(axis cs:7,0.290340803768487)
--(axis cs:6,0.336765392604221)
--(axis cs:5,0.363441497262834)
--(axis cs:4,0.405154174527043)
--(axis cs:3,0.528182538832653)
--(axis cs:2,0.664963657108469)
--(axis cs:1,0.978085393178987)
--cycle;

\path [draw=white, fill=color1, opacity=0.3]
(axis cs:1,0.98006022)
--(axis cs:1,0.98040974)
--(axis cs:2,0.97884724)
--(axis cs:3,0.97697979)
--(axis cs:4,0.97466717)
--(axis cs:5,0.97225095)
--(axis cs:6,0.9693497)
--(axis cs:7,0.96475306)
--(axis cs:8,0.95872132)
--(axis cs:9,0.95026569)
--(axis cs:10,0.93879402)
--(axis cs:10,0.93444794)
--(axis cs:10,0.93444794)
--(axis cs:9,0.94726093)
--(axis cs:8,0.95652238)
--(axis cs:7,0.96375546)
--(axis cs:6,0.96859526)
--(axis cs:5,0.97161101)
--(axis cs:4,0.97414029)
--(axis cs:3,0.97661017)
--(axis cs:2,0.9784977)
--(axis cs:1,0.98006022)
--cycle;

\path [draw=white, fill=color2, opacity=0.3]
(axis cs:1,0.976215192339312)
--(axis cs:1,0.977384807660689)
--(axis cs:2,0.965908675825942)
--(axis cs:3,0.954396550293454)
--(axis cs:4,0.938833594467194)
--(axis cs:5,0.915414053798822)
--(axis cs:6,0.895619685640007)
--(axis cs:7,0.884494106362243)
--(axis cs:8,0.864107510531119)
--(axis cs:9,0.850402209180996)
--(axis cs:10,0.828300190572827)
--(axis cs:10,0.815863809427173)
--(axis cs:10,0.815863809427173)
--(axis cs:9,0.839988901930115)
--(axis cs:8,0.857282489468881)
--(axis cs:7,0.877917322209185)
--(axis cs:6,0.892580314359993)
--(axis cs:5,0.908201946201178)
--(axis cs:4,0.932876405532806)
--(axis cs:3,0.948763449706546)
--(axis cs:2,0.963191324174058)
--(axis cs:1,0.976215192339312)
--cycle;

\path [draw=white, fill=color3, opacity=0.3]
(axis cs:1,0.976715192339311)
--(axis cs:1,0.977884807660689)
--(axis cs:2,0.9758353459697)
--(axis cs:3,0.974710737364271)
--(axis cs:4,0.973421617008257)
--(axis cs:5,0.972147745543039)
--(axis cs:6,0.970689647256434)
--(axis cs:7,0.969317268245082)
--(axis cs:8,0.967497388500073)
--(axis cs:9,0.965735620757116)
--(axis cs:10,0.963105855422206)
--(axis cs:10,0.956950144577794)
--(axis cs:10,0.956950144577794)
--(axis cs:9,0.964788823687329)
--(axis cs:8,0.966572611499927)
--(axis cs:7,0.968568446040632)
--(axis cs:6,0.969997019410233)
--(axis cs:5,0.971236254456961)
--(axis cs:4,0.972778382991743)
--(axis cs:3,0.974182595969062)
--(axis cs:2,0.9751846540303)
--(axis cs:1,0.976715192339311)
--cycle;

\addplot [semithick, color0, mark=*, mark size=2, mark options={solid}]
table {%
1 0.97888
2 0.67821
3 0.538706666666667
4 0.415535
5 0.371068
6 0.339676666666667
7 0.295862857142857
8 0.2617
9 0.259822222222222
10 0.23938
};
\addlegendentry{MAP}

\addplot [semithick, color1, mark=*, mark size=2, mark options={solid}]
table {%
1 0.98023498
2 0.97867247
3 0.97679498
4 0.97440373
5 0.97193098
6 0.96897248
7 0.96425426
8 0.95762185
9 0.94876331
10 0.93662098
};
\addlegendentry{VB (VOGN)}

\addplot [semithick, color2, mark=*, mark size=2, mark options={solid}]
table {%
1 0.9768
2 0.96455
3 0.95158
4 0.935855
5 0.911808
6 0.8941
7 0.881205714285714
8 0.860695
9 0.845195555555555
10 0.822082
};
\addlegendentry{LA-Diag}

\addplot [semithick, color3, mark=*, mark size=2, mark options={solid}]
table {%
1 0.9773
2 0.97551
3 0.974446666666667
4 0.9731
5 0.971692
6 0.970343333333333
7 0.968942857142857
8 0.967035
9 0.965262222222222
10 0.960028
};
\addlegendentry{LA-KFAC}

\end{axis}

\end{tikzpicture}
    \vspace{-0.35em}

    \caption{Continual learning results on Permuted-MNIST. MAP fails catastrophically as more tasks are added. The Bayesian approaches substantially outperform MAP, with LA-KFAC performing the best, closely followed by VOGN.}
    \label{fig:continual_learning}
\end{wrapfigure}

Beyond predictive uncertainty quantification, the LA is useful in wide range of applications such as Bayesian optimization \citep{snoek2015scalable}, bandits \citep{chapelle2011empirical}, active learning \citep{mackay1992evidence,park2011active}%
, and continual learning \citep{ritter2018online}.
The \libname{} library conveniently facilitates these applications.
As an example, we demonstrate the performance of the LA on the standard continual learning benchmark with the Permuted-MNIST dataset, consisting of ten tasks each containing pixel-permuted MNIST images \citep{goodfellow2013empirical}.
\Cref{fig:continual_learning} shows how the all-layer diagonal and Kronecker-factored LAs can overcome \emph{catastrophic forgetting}.
In this experiment, we update the LAs after each task as suggested by \citet{ritter2018online} and improve upon their result by tuning the prior precision through marginal likelihood optimization during training, following \citet{immer2021scalable}
(details in \cref{app:sec:cl_details}).
Using this scheme, the performance after $10$ tasks is at around $96\%$ accuracy, outperforming other Bayesian approaches for continual learning~\citep{nguyen2018variational, titsias2019functional, pan2020continual}. Concretely, we show that the KFAC LA, while much simpler when applied via \libname, can achieve better performance to a recent VB baseline \citep[VOGN,][]{osawa2019practical}.
Our library thus provides an easy and quick way of constructing a strong baseline for this application.

    \section{Related Work}
    \label{sec:related_work}
    The LA is fundamentally a local approximation that covers a single mode of the posterior; similarly, other Gaussian approximations such as mean-field variational inference \citep{graves2011practicalVI,blundell2015weight,osawa2019practical} or SWAG \citep{maddox2019simple} also only capture local information. %
SWAG uses the first and second empirical moment of SGD iterates to form a diagonal plus low-rank Gaussian approximation but requires storing many NN copies and applying a (costly) heuristic related to batch normalization at test time. In contrast, the LA directly uses curvature information of the loss around the MAP and can be applied \emph{post-hoc} to pre-trained~NNs.

In contrast to local Gaussian approximations, (stochastic-gradient) MCMC methods \citep[etc.]{welling2011bayesian,wenzel2020good,zhang2020csgmcmc,izmailov2021bayesian,garriga2021exact} and deep ensembles \citep{lakshminarayanan2017simple} can explore several modes. Nevertheless, prior works---also validated in our experiments in \Cref{sec:experiments}---indicate that using a single mode might not be as limiting in practice as one might think.
\citet{wilson2020bayesian} conjecture that this is due to the complex, nonlinear connection between the parameter space and the function (output) space of NNs.
Moreover, while unbiased compared to its simpler alternatives, MCMC methods are notoriously expensive in practice and, thus, often require further approximations such as distillation \citep{korattikara2015bayesian,wang2018adversarial}.
Finally, note that both the LA as well as SWAG can be extended to ensembles of modes in a \emph{post-hoc} manner \citep{eschenhagen2021mola,wilson2020bayesian}.

    \section{Conclusion}
    \label{sec:conclusion}
    
In this paper, we argued that the Laplace approximation is a simple yet %
competitive and versatile method for Bayesian deep learning that deserves wider adoption. To this end, we reviewed many recent advances to and variants of the Laplace approximation, including versions with minimal cost overhead that can be applied \emph{post-hoc} to pre-trained off-the-shelf models.
In a comprehensive evaluation we demonstrated that the Laplace approximation is on par with other approaches that approximate the intractable network posterior, but at typically much lower computational cost. A particularly simple variant that only treats some weights probabilistically can even be used in the context of pre-trained transformer models to improve predictive uncertainty.
As an efficient implementation is not straightforward, we introduced \libname{}, a modular and extensible software library for PyTorch offering user-friendly access to all major flavors of the Laplace approximation. In this way, Laplace approximations provide drop-in Bayesian functionality for most types of deep neural networks.

    \begin{ack}
        We thank Kazuki Osawa for providing early access to his automatic second-order differentiation (ASDL) library for PyTorch, Alex Botev for feedback on our manuscript, and Simo Ryu for spotting a sign typo in \cref{eq:laplace_approx}.
        We also thank the anonymous reviewers for their helpful suggestions.
        
        E.D. acknowledges funding from the EPSRC and Qualcomm.
        A.I. gratefully acknowledges funding by the Max Planck ETH Center for Learning Systems (CLS).
        R.E., A.K. and P.H. gratefully acknowledge financial support by the European Research Council through ERC StG Action 757275 / PANAMA; the DFG Cluster of Excellence “Machine Learning - New Perspectives for Science”, EXC 2064/1, project number 390727645; the German Federal Ministry of Education and Research (BMBF) through the Tübingen AI Center (FKZ: 01IS18039A); and funds from the Ministry of Science, Research and Arts of the State of Baden-Württemberg.
        A.K. is grateful to the International Max Planck Research School for Intelligent Systems (IMPRS-IS) for support.
    \end{ack}

    {
        \small
        \typeout{}
        \bibliography{main}
        \bibliographystyle{unsrtnat}
    }

    \clearpage

    \begin{appendices}
        \setcounter{equation}{0}
        \crefalias{section}{appsec}
        \crefalias{subsection}{appsec}

        \section{Derivation}
        \label{app:sec:add_details}
        
\subsection{The Derivation of the Laplace Approximation}

Let $p(\theta \mid \D)$ be an intractable posterior, written as
\begin{equation} \label{eq:posterior_appendix}
    p(\theta \mid \D) := \frac{1}{\int p(\D \mid \theta) p(\theta) \,d\theta} p(\D \mid \theta) p(\theta) =: \frac{1}{Z} h(\theta)
\end{equation}
Our goal is to approximate this distribution with a Gaussian arising from the Laplace approximation. The key observation is that we can rewrite the normalizing constant $Z$ as the integral $\int \exp(\log h(\theta)) \,d\theta$. Let $\theta_\map := \argmax_\theta \log p(\theta \mid \D) = \argmax_\theta \log h(\theta)$ be a (local) maximum of the posterior---the so-called \defword{maximum a posteriori (MAP)} estimate. Taylor-expanding $\log h$ around $\theta_\map$ up to the second order yields
\begin{equation} \label{eq:log_posterior_taylor}
    \log h(\theta) \approx h(\theta_\map) - \frac{1}{2} (\theta - \theta_\map)^\top \varLambda \, (\theta - \theta_\map) ,
\end{equation}
where $\varLambda := -\nabla^2 \log h(\theta) \vert_{\theta_\map}$ is the negative Hessian matrix of the log-joint in \eqref{eq:posterior_appendix}, evaluated at $\theta_\map$. Similar to its original formulation, here we again obtain a (multivariate) Gaussian integral, the analytic solution of which is readily available:
\begin{equation} \label{eq:evidence_laplace}
    \begin{aligned}
    Z &\approx \exp(\log h(\theta_\map)) \int \exp \left(-\frac{1}{2} (\theta - \theta_\map)^\top \varLambda \, (\theta - \theta_\map) \right) \,d\theta \\
        &= h(\theta_\map) \frac{(2 \pi)^\frac{d}{2}}{(\det \varLambda)^\frac{1}{2}} .
    \end{aligned}
\end{equation}
Plugging the approximations \eqref{eq:log_posterior_taylor} and \eqref{eq:evidence_laplace} back into the expression of $p(\theta \mid \D)$, we obtain
\begin{equation} \label{eq:laplace_posterior}
    p(\theta \mid \D) = \frac{1}{Z} h(\theta) \approx \frac{(\det \varLambda)^\frac{1}{2}}{(2 \pi)^\frac{d}{2}} \exp \left(- \frac{1}{2} (\theta - \theta_\map)^\top \varLambda \, (\theta - \theta_\map) \right) ,
\end{equation}
which we can immediately identify as the Gaussian density $\N(\theta \mid \theta_\map, \varSigma)$ with mean $\theta_\map$ and covariance matrix $\varSigma := \varLambda^\inv$.

        \section{Details on the Four Components}
        \label{app:sec:components}
        
\subsection*{\protect\circled{inner sep=1pt}{\small 1}\hspace{0.5em} Inference over Subsets of Weights}
\setcounter{subsection}{1}

\subsubsection{Subnetwork}
Storing the full $D \times D$ covariance matrix $\varSigma$ of the weight posterior in \cref{eq:laplace_posterior} is computationally intractable for a modern neural networks. 
One approach to reduce this computational burden is to perform inference over only a small \emph{subset} of the model parameters $\theta$ \citep{daxberger2020expressive}.
This is motivated by recent findings that neural nets can be heavily pruned without sacrificing test accuracy \citep{frankle2018the}, and that in the neighborhood of a local optimum, there are many directions that leave the predictions unchanged \citep{maddox2020rethinking}.

This \emph{subnetwork inference} approach uses the following approximation to the posterior in \cref{eq:laplace_posterior}:
\begin{align}
    \label{eq:subnet_posterior}
     p(\theta \mid \D) \ \ \approx \ \ p(\theta_S \mid \D) \ \prod_{r} \delta(\theta_{r} - \widehat{\theta}_{r} ) \ \ = \ \ q_S(\theta) \ ,
\end{align}
where $\delta(x - a)$ denotes the Dirac delta function centered at $a$.
The approximation $q_S(\theta)$ in \cref{eq:subnet_posterior} simply decomposes the full neural network posterior $p(\theta \mid \D)$ into a Laplace posterior $p(\theta_S \mid\D)$ over the subnetwork $\theta_S \in \mathbb{R}^S$, and fixed, deterministic values $\widehat{\theta}_{r}$ to the $D-S$ remaining weights $\theta_{r}$.
In practice, the remaining weights $\theta_{r}$ are simply set to their MAP estimates, i.e.\ $\widehat{\theta}_{r} = \theta_{r}^{\text{MAP}}$, requiring no additional computation.
Importantly, note that the subnetwork size $S$ is in practice a hyperparameter that can be controlled by the user.
Typically, $S$ will be set such that the subnetwork is much smaller than the full network, i.e. $S \ll D$.
In particular, $S$ can be set such that it is tractable to compute and store the full $S \times S$ covariance matrix over the subnetwork.
This allows us to capture rich dependencies across the weights within the subnetwork.
However, in principle one could also employ one of the (less expressive) factorizations of the Hessian/Fisher described in \cref{app:sec:factorization}.

\citet{daxberger2020expressive} propose to choose the subnetwork such that the subnetwork posterior $q_S(\theta)$ in \cref{eq:subnet_posterior} is as close as possible (w.r.t.\ some discrepancy measure) to the full posterior $p(\theta \mid \D)$ in \cref{eq:laplace_posterior}. 
As the subnetwork posterior is degenerate due to the involved Dirac delta functions, common discrepancy measures such as the KL divergence are not well defined.
Therefore, \citet{daxberger2020expressive} propose to use the squared 2-Wasserstein distance, which in this case takes the following form:
\begin{equation}
    \label{eq:wass2squared}
    W_2(p(\theta \mid \D), q_{S}(\theta))^2 = \text{Tr}\left(\varSigma + \varSigma_{S} - 2 \left(\varSigma_{S}^{1/2} \ \varSigma \ \varSigma_{S}^{1/2}\right)^{1/2} \right)\ , 
\end{equation}
where the (degenerate) subnetwork covariance matrix $\varSigma_{S}$ is equal to the full covariance matrix $\varSigma$ but with zeros at the positions corresponding to the weights $\theta_r$ (i.e.\ those \emph{not} part of the subnetwork).

Unfortunately, finding the subset of weights $\theta_S \in \mathbb{R}^S$ of size $S$ that minimizes \cref{eq:wass2squared} is combinatorially hard, as the contribution of each weight depends on every other weight. 
\citet{daxberger2020expressive} therefore assume that the weights are independent, resulting in the following simplified objective:
\begin{equation}
    \label{eq:wass2squared_indep}
    W_2(p(\theta \mid \D), q_{S}(\theta))^2 
    \approx \sum_{d=1}^{D} \sigma^2_d(1 - m_d)\ , 
\end{equation}
where $\sigma^2_d = \varSigma_{dd}$ is the marginal variance of the $d^{\text{th}}$ weight, and $m_d = 1$ if $\theta_d\,\in\,\theta_S$ (with slight abuse of notation) or 0 otherwise is a binary mask indicating which weights are part of the subnetwork (see \citet{daxberger2020expressive} for details).
The objective in \cref{eq:wass2squared_indep} is trivially minimized by choosing a subnetwork containing the $S$ weights with the highest $\sigma^2_d$ values (i.e.\ with largest marginal variances).

In practice, even computing the marginal variances (i.e.\ the diagonal of $\varSigma$) is intractable, as it requires storing and inverting the Hessian/Fisher $\varLambda$.
To approximate the marginal variances, one could use a diagonal Laplace approximation \citep{denker1990transforming,kirkpatrick2017overcoming} that assumes $\text{diag}(\varSigma) \approx \text{diag}(\varLambda)^{-1}$.
Alternatively, one could use diagonal SWAG \citep{maddox2019simple}.
For more details on subnetwork inference, refer to~\citet{daxberger2020expressive}.

\subsubsection{Last-Layer}

The last-layer Laplace \citep{snoek2015scalable,kristiadi2020being} is a special variant of the subnetwork Laplace where $\theta_S$ in \eqref{eq:subnet_posterior} is assumed to equal the last-layer weight matrix $W^{(L)}$ of the network. That is, we let $f_\theta: \R^M \to \R^C$ is an $L$-layer NN, and assume that the first $L-1$ layers of $f_\theta$ is a feature map. Given MAP-trained parameters $\theta_\map$, we define a Laplace-approximated posterior over $W^{(L)}$
\begin{equation}
    p(W^{(L)} \mid \D) \approx \N(W^{(L)} \mid W^{(L)}_\map, \varSigma^{(L)}) ,
\end{equation}
and we leave the rest of the parameters with their MAP-estimated values. Since this matrix is small relative to the entire network, the last-layer Laplace can be implemented efficiently.

\subsection*{\protect\circled{inner sep=1pt}{\small 2}\hspace{0.5em} Hessian Factorization}
\label{app:sec:factorization}

For brevity, given a datum $(x, y)$, we denote $s(x, y)$ to be the gradient of the log-likelihood at $\theta_\map$, i.e.
\begin{equation*}
    s(x, y) := \nabla_\theta p(y \mid f_\theta(x)) \vert_{\theta_\map} .
\end{equation*}
Using this notation, we can write the Fisher compactly by 
\begin{equation} \label{eq:FIM_appendix}
    F := \textstyle\sum_{n=1}^N \E_{p(y \mid f_\theta(x_n))} \left( s(x_n, y) s(x_n, y)^\intercal \right) ,
\end{equation}
We shall refer to this matrix as the \emph{full Fisher}. Recall that $F$ is as large as the exact Hessian of the network, so its computation is often infeasible. Thus, here, we review several factorization schemes that makes the computation (and storage) of the Fisher efficient, starting from the simplest.

\paragraph{Diagonal}
Although MacKay recommended to not use the diagonal factorization of the Hessian \citep{mackay1992practical}, a recent work has indicated this factorization is usable for sufficiently deep NNs \citep{farquhar2020liberty}. In this factorization, we simply assume that the negative-log-posterior's Hessian $\varLambda$ is simply a diagonal matrix with diagonal elements equal the diagonal of the Fisher, i.e. $\varLambda \approx -\diag{F}^\top I - \lambda I$. Since we can write $\diag{F} = \sum_{n=1}^N \E_{p \left(y \mid f_{\theta_\map}(x_n)\right)} (s(x_n, y) \odot s(x_n, y))$,\footnote{The operator $\odot$ denotes the Hadamard product.} this factorization is efficient: Not only does it require only a vector of length $D$ to represent $F$ but also it incurs only a $O(D)$ cost when inverting $\varLambda$---down from $O(D^3)$.

\paragraph{KFAC}
The KFAC factorization can be seen as a midpoint between the two extremes: diagonal factorization, which might be too restrictive, and the full Fisher, which is computationally infeasible. The key idea is to model the correlation between weights in the same layer but assume that any pair of weights from two different layers are independent---this is a more sophisticated assumption compared to the diagonal factorization since there, it is assumed that \emph{all} weights are independent of each other. For any layer $l = 1, \dots, L$, denoting $N_l$ as the number of hidden units at the $l$-th layer, let $W^{(l)} \in \R^{N_{l} \times N_{l-1}}$ be the weight matrix of the $l$-th layer of the network, $a^{(l)}$ the $l$-th hidden vector, and $g^{(l)} \in \R^{N_{l}}$ the log-likelihood gradient w.r.t. $a^{(l)}$. For each $l = 1, \dots, L$, we can then write the outer product inside expectation in \eqref{eq:FIM} as $s(x_i, y) s(x_i, y)^\top = a^{(l-1)} a^{(l)}{}^\top \otimes g^{(l)} g^{(l)}{}^\top$. Furthermore, assuming that $a^{(l-1)}$ is independent of $g^{(l)}$, we obtain the approximation of the $l$-th diagonal block of $F$, which we denote by $F^{(l)}$:
\begin{equation} \label{eq:FIM_KF}
    F^{(l)} \approx \E\left( a^{(l-1)} a^{(l-1)}{}^\top \right) \otimes \E\left( g^{(l)} g^{(l)}{}^\top \right) =: A^{(l-1)} \otimes G^{(l)} ,
\end{equation}
where we represent both the sum and the expectation in \eqref{eq:FIM_appendix} as $\E$ for brevity.

From the previous expression we can see that the space complexity for storing $F^{(l)}$ is reduced to $O(N_{l}^2 + N_{l-1}^2)$, down from $O(N_{l}^2 N_{l-1}^2)$. Considering all $L$ layers of the network, we obtain the layer-wise Kronecker factors $\{ A^{(l)} \}_{l = 0}^{L-1}$ and $\{ G^{(l)} \}_{l = 1}^L$ of the log-likelihood's Hessian. This corresponds to the block-diagonal approximation of the full Hessian.

One can then readily use these Kronecker factors in a Laplace approximation. For each layer $l$, we obtain the $l$-th diagonal block of $\varLambda$---denoted $\varLambda^{(l)}$---by
\begin{equation*}
    \begin{aligned}
        \varLambda^{(l)} &\approx \left( A^{(l-1)} + \sqrt{\lambda} I \right) \otimes \left( G^{(l)} + \sqrt{\lambda} I \right) \\
            &=: V^{(l)} \otimes U^{(l)} .
    \end{aligned}
\end{equation*}
Note that we take the square root of the prior precision to avoid ``double-counting'' the effect of the prior. Nonetheless, this can still be a crude approximation~\citep{martens2015optimizing, immer2020improving}. This particular Laplace approximation has been studied by \citet{ritter2018scalable,ritter2018online} and can be seen as approximating the posterior of each $W^{(l)}$ with the matrix-variate Gaussian distribution \citep{gupta1999matrix}: $p(W^{(l)} \mid \D) \approx \MN(W^{(l)} \mid W^{(l)}_\map, U^{(l)}{}^\inv, V^{(l)}{}^\inv)$. Hence, sampling can be done easily in a layer-wise manner:
\begin{align*}
    W^{(l)} &\sim p\left(W^{(l)} \mid \D\right) \iff W^{(l)} = W^{(l)}_\map + U^{(l)}{}^{-\frac{1}{2}} E V^{(l)}{}^{-\frac{1}{2}} \\
    \intertext{where}
    \quad E &\sim \MN(0, I_{N_{l}}, I_{N_{l-1}}) ,
\end{align*}
where we have denoted by $I_b$ the identity $b \times b$ matrix, for $b \in \mathbb{N}$. Note that the above matrix inversions and square-root are in general much cheaper than those involving the entire $\varLambda$. Sampling $E$ is not a problem either since $\MN(0, I_{N_{l}}, I_{N_{l-1}})$ is equivalent to the standard $(N_{l} N_{l-1})$-variate Normal distribution.
As an alternative, \citet{immer2020improving} suggest to incorporate the prior exactly using an eigendecomposition of the individual Kronecker factors, which can improve performance.

\paragraph{Low-rank block-diagonal} We can improve KFAC's efficiency by considering its low-rank factorization \citep{lee2020estimating}. The key idea is to eigendecompose the Kronecker factors in \eqref{eq:FIM_KF} and keep only the eigenvectors corresponding to the first $k$ largest eigenvalues. This can be done employing the eigenvalue-corrected KFAC \citep{george2018fast}. That is, for each layer $l = 1, \dots, L$:
\begin{equation*}
    \begin{aligned}
        F^{(l)} &\approx \left(U_A^{(l-1)} S_A^{(l-1)} U_A^{(l-1)}{}^\top\right) \otimes \left(U_G^{(l)} S_G^{l} U_G^{(l)}{}^\top\right) \\
            &= \left( U_A^{(l-1)} \otimes U_G^{(l)} \right) \left( S_A^{(l-1)} \otimes S_G^{(l)} \right) \left( U_A^{(l-1)} \otimes U_G^{(l)} \right)^\top .
    \end{aligned}
\end{equation*}
Under this decomposition, one can the easily obtain the optimal rank-$k$ approximation of $F^{(l)}$, denoted by $F^{(l)}_k$, by selecting the top-$k$ eigenvalues. However, the diagonal of this rank-$k$ matrix can deviate too far from the exact diagonal elements of $F^{(l)}$. Hence, one can make the diagonal of this low rank matrix exact replacing $\diag{F^{l}_k}$ with $\diag{F^{(l)}}$, and obtain the following rank-$k$-plus-diagonal approximation of $F^{(l)}$:
\begin{equation*}
    F^{(l)} \approx F^{(l)}_k + \diag{F^{(l)}} - \diag{F^{(l)}_k} .
\end{equation*}
This factorization can be seen as a combination of the previous two approximations: For each diagonal block of $F$, we use the exact diagonal elements of $F$ and approximate the off-diagonal elements with a rank-$k$ matrix arising from KFAC. Both the space and computational complexities are lower than those of KFAC since here we work exclusively with truncated and diagonal matrices.

\paragraph{Low-rank} Instead of only approximating each block by a low-rank structure, the entire Hessian or GGN can also be approximated by a low-rank structure~\citep{sharma2021sketching, maddox2020rethinking}.
Eigendecomposition of $F$ is a convenient way to obtain a low-rank approximation.
The eigendecomposition of $F$ is given by $Q L Q^\top$ where the columns of $Q\in \mathbb{R}^{D\times D}$ are eigenvectors of $F$ and $L=\diag{l}$ is a $D$-dimensional diagonal matrix of eigenvalues.
Assuming the eigenvalues in $l$ are arranged in a descending order, the optimal $k$-rank approximation in Frobenius or spectral norm is given by truncation~\citep{eckart1936approximation}:
let $\widehat{Q} \in \mathbb{R}^{D \times k}$ be the matrix of the first $k$ eigenvectors corresponding to the largest $k$ eigenvalues $\widehat{l} \in \mathbb{R}^k$.
That is, we truncate all eigenvectors and eigenvalues after the $k$ largest eigenvalues.
The low-rank approximation is then given by
\begin{equation*}
    F \approx \widehat{Q} \, \diag{\widehat{l}} \, \widehat{Q}^\top.
\end{equation*}
The rank $k$ can be chosen based on the eigenvalues so as to retain as much information of the Hessian (approximation) as possible.
Further, sampling and computation of the log-determinant can be carried out efficiently.

\paragraph{Functional}
When considering network linearization for the predictive distribution, we can directly infer the Gaussian distribution on the outputs, of which there are typically few, instead of inferring a distribution on the parameters, of which there are many~\citep{khan2019approximate, immer2020improving}.

\subsection*{\protect\circled{inner sep=1pt}{\small 3}\hspace{0.5em} Hyperparameter Tuning}

In this section we focus on tuning the prior variance/precision hyperparameter for simplicity. The same principle can be used for other hyperparameters of the Laplace approximation such that observation noise in the case of regression.

\paragraph{\emph{Post-Hoc}} Here, we assume that the steps of the Laplace approximation---MAP training and forming the Gaussian approximation---as two independent steps. As such, we are free to choose different prior variance $\gamma^2$ in the latter part, irrespective to the weight decay hyperparameter used in the former. Here, we review several ways to optimize $\gamma^2$ \emph{post-hoc}. \citet{ritter2018scalable} proposes to tune $\gamma^2$ by maximizing the posterior-predictive over a validation set $\D_\text{val} := ( x_n, y_n )_{n=1}^{N_\text{val}}$. That is we solve the following one-parameter optimization problem:
\begin{equation} \label{eq:tuning_loglik}
    \gamma^2_* = \argmax_{\gamma^2} \sum_{n=1}^{N_\text{val}} \log p(y_n \mid x_n, \D) .
\end{equation}
However, \citet{kristiadi2020being} found that the previous objective tends to make the Laplace approximation overconfident to outliers. Hence, they proposed to add an auxiliary term that depends on an OOD dataset $\D_\text{out} := (x^{(\text{out})}_n)_{n=1}^{N_\text{out}}$ to \eqref{eq:tuning_loglik}, as follows
\begin{equation}
    \gamma^2_* = \argmax_{\gamma^2} \sum_{n=1}^{N_\text{val}} \log p(y_n \mid x_n, \D) + \lambda \sum_{n=1}^{N_\text{out}} H \left[ p(y_n \mid x^{(\text{out})}_n, \D) \right] ,
\end{equation}
where $H$ is the entropy functional and $\lambda \in (0, 1]$ is a trade-off hyperparameter. Intuitively, we choose $\gamma^2$ that balances the calibration on the true dataset and the low-confidence on outliers. 
Moreover, other losses could be constructed to tune the prior precision for optimal performance w.r.t. some desired quantity.
Finally, inspired by \citet{immer2021scalable} (further details below in \emph{Online}) one can also maximize the Laplace-approximated marginal likelihood \eqref{eq:evidence_laplace} to obtain $\gamma^2_*$, which eliminates the need for the validation data.

\paragraph{\emph{Online}} Contrary to the \emph{post-hoc} tuning above, here we perform a Laplace approximation and tune the prior variance simultaneously as we perform a MAP training \citep{immer2021scalable}. The key is to form a Laplace-approximated posterior every $B$ epochs of a gradient descent, and use this posterior to approximate the marginal likelihood, cf. \eqref{eq:evidence_laplace}. By maximizing this marginal likelihood, we can find the best hyperparameters. Thus, once the MAP training has finished, we automatically obtain a prior variance that is already suitable for the Laplace approximation. Note that, this way, only a single MAP training needs to be done. This is in contrast to the classic, offline evidence framework \citep{mackay1992evidence} where the marginal likelihood maximization is performed only when the MAP estimation is done, and these steps need to be iteratively done until convergence. As a final note, similar to the \emph{post-hoc} marginal likelihood above, this \emph{online Laplace} does not require a validation set and has an additional benefit of improving the network's generalization performance \citep{immer2021scalable}. We refer the reader to \cref{alg:online_laplace} for an overview.

\begin{algorithm}[t]
    \caption{Online Laplace (adapted from \citet[Algorithm 1]{immer2021scalable})}
    \label{alg:online_laplace}
    \begin{algorithmic}[1]
        \Require
            \Statex NN $f_\theta$; training set $\D$; learning rate $\alpha_0$ and number of epochs $T_0$ for MAP estimation; learning rate $\alpha_1$ and number of epochs $T_1$ for hyperparameter tuning; marginal likelihood maximization frequency $F$.

        \vspace{1em}

        \State Initialize $\theta_0$
        \For{$t = 1, \dots, T_0$}
            \State $g_t \leftarrow \nabla_\theta \L(\D; \theta) \vert_{\theta_{t-1}}$
            \State $\theta_t \leftarrow \theta_{t-1} - \alpha_0 \, g_t$
            \If{$t \mod F = 0$}
                \State $p(\theta \mid \D) \approx \N(\theta \mid \theta_t, (\nabla^2 \L(\D; \theta) \vert_{\theta_t})^\inv)$  \Comment{Perform a Laplace approximation}
                \For{$\widetilde{t} = 1, \dots, T_1$}  \Comment{Hyperparameter optimization}
                    \State $h_{\widetilde{t}} \leftarrow \nabla_{\gamma^2} \log p(\D \mid \gamma^2) \vert_{\gamma^2_{\widetilde{t}-1}}$    \Comment{The marginal likelihood follows from \eqref{eq:evidence_laplace}}
                    \State $\gamma^2_{\widetilde{t}} \leftarrow \gamma^2_{\widetilde{t}-1} + \alpha_1 \, h_{\widetilde{t}}$
                \EndFor
            \EndIf
        \EndFor
        \State \Return $\theta_{T_0}$; $\nabla^2 \L(\D; \theta) \vert_{\theta_{T_0}}$
    \end{algorithmic}
\end{algorithm}

\subsection*{\protect\circled{inner sep=1pt}{\small 4}\hspace{0.5em} Approximate Predictive Distribution}
\setcounter{subsection}{4}

Here, we denote $x_* \in \R^N$ to be a test point, and $f_*$ be the network output at this point. We will review different way to approximate the predictive distribution $p(y \mid x_*, \D)$ given a Gaussian approximate posterior, starting from the most general.

\subsubsection{General}

\vspace{-0.5em}
\paragraph*{Monte Carlo Integration} The simplest but general and unbiased approximation is the Monte Carlo (MC) integration, which can be performed by sampling an approximate posterior $q(\theta \mid \D)$ repeatedly: 
\begin{equation*}
    p(y \mid x_*, \D) \approx \frac{1}{S} \sum_{s=1}^S p(y \mid f_{\theta_s}(x_*)) ,  \qquad \text{where } \theta_s \sim q(\theta \mid \D) .
\end{equation*}
While the error of this approximation decays like $1/\sqrt{S}$ and thus requires many samples to be accurate, for practical BNNs, it is standard to use 10 or 20 samples of $q(\theta \mid \D)$ \citep[etc.]{ritter2018scalable,kristiadi2020being,blundell2015weight}. Note that this approximation can be used regardless the form of the likelihood $p(y \mid f_{\theta}(x))$, in particular it can be used to directly obtain the predictive distribution in both the regression and classification alike.

\subsubsection{Distribution of Network Outputs}

Here, we are concerned in approximating the marginal distribution of $f(x_*)$, where $\theta$ has been integrated out.

\paragraph*{Linearization} In this approximation, we linearize the network to obtain 
\begin{equation*}
    f_\theta(x_*) \approx f_{\theta_\map}(x_*) + J_*^\top (\theta - \theta_\map) ,
\end{equation*}
where $J_* := \nabla_\theta f_\theta(x_*) |_{\theta_\map} \in \R^{d \times c}$ is the Jacobian matrix of the network output. This way, under a Gaussian approximate posterior $q(\theta \mid \D)$, the marginal distribution over the network output $f_* := f(x_*)$ is again a Gaussian, given by\footnote{See \citet[Sec. 4.5.2]{bishop2006prml}.} 
\begin{equation*}
    \begin{aligned}
        p(f_* \mid f_\theta(x_*), x_*, \D) &= \int \delta(f_* - f_{\theta}(x_*)) \, q(\theta \mid \D) \,d\theta \\ 
            &\approx \N(f_* \mid f_{\theta_\map}(x_*), J_*^\top \varSigma J_*)  
    \end{aligned}
\end{equation*}
This approximation has been extensively used for small networks \citep{mackay1992evidence}, but it has since gone out of favor in deep learning due to its cost---the Jacobian $J_*$ needs to be computed \emph{per input point}. Nevertheless, this approximation is still useful in theoretical works due to its analytical nature \citep{kristiadi2020being,kristiadi2020learnable,eschenhagen2021mola} . Moreover, in problems where it can be efficiently use in practice, it offers a better approximation than MC-integral \citep{immer2020improving,foong2019between}.
Due to the linearization in the network parameters, it is further possible to obtain a functional prior in the form of a Gaussian process~\citep{khan2019approximate, immer2020improving}.
This allows to perform function-space inference as opposed to weight-space inference which is amenable to different Hessian approximations than those pointed out above in \cref{app:sec:factorization}, and is, for example, useful for continual learning~\citep{pan2020continual}.

\subsubsection{Regression}

Assume that we already have a Gaussian approximation to $p(f_* \mid x_*, \D) \approx \N(f_* \mid \mu_*, \varSigma_*)$ via the linearization above. In regression, we still need to incorporate the observation noise $\beta$ encoded in the (usually) Gaussian likelihood $\N(y_* \mid f_*, \beta I)$\footnote{We assume a multivariate output $y_* \in \R^C$ for full generality.} to make prediction. This can be easily done in an exact manner:
\begin{equation*}
    \begin{aligned}
        p(y_* \mid x_*) &= \int_{\R^C} \N(y_* \mid f_*, \beta I) \, \N(f_* \mid \mu_*, \varSigma_*) \,df_* \\
            &= \N(y_* \mid \mu_*, \varSigma_* + \beta I) ,
    \end{aligned}
\end{equation*}
since the integral above is just a convolution of two Gaussian r.v.s.

\subsubsection{Classification and Generalized Regression}

Since unlike the regression case, the classification likelihood $p(y_* \mid f_*)$ is non-Gaussian, we cannot analytically obtain $p(y_* \mid x_*)$ given a Gaussian approximation $p(f_* \mid x_*, \D) \approx \N(f_* \mid \mu_*, \varSigma_*)$. So, in this case we are interested in approximating the intractable integral 
\begin{equation*}
    p(y_* \mid x_*) = \int p(y_* \mid f_*) \, \N(f_* \mid \mu_*, \varSigma_*) \,df_* ,
\end{equation*}
where $p(y_* \mid f_*)$ is constructed via an inverse-link function. Here we will review the usual case of classification, i.e. when $p(y_* \mid f_*) = \sigma(f_*)$ where $\sigma$ is the logistic-sigmoid function, or $p(y_* \mid f_*) = \softmax(f_*)$ .

\paragraph*{Delta Method} The crux of the delta method \citep{wu2018deterministic,ahmed2007seeking,braun2010variational} is a Taylor-expansion of the softmax function around $\mu_*$ up to the second order. Then, since $p(f_* \mid x_*, \D)$ is assumed to be Gaussian, the integral $\E_{p(f_* \mid x_*, \D)} (\softmax(f_*))$ can be computed easily, resulting in an analytic expression $\softmax(\mu_*) + 1/2 \, \mathrm{tr}(B\varSigma_*)$, where $B$ is the Hessian matrix of the softmax at $\mu_*$.

\paragraph*{Probit Approximations} The essence of the (binary) probit approximation \citep{spiegelhalter1990sequential,mackay1992evidence} is to approximate $\sigma$ with the probit function $\Phi$---the standard Normal c.d.f.---which makes the integral solvable analytically. Using this approximation, one can then obtain the closed-form approximation
\begin{equation*}
    \begin{aligned}
        p(y_* \mid x_*) &\approx \int_\R \Phi(f_*) \, \N(f_* \mid \mu_*, \sigma^2_*) \,df_* \\
            &= \sigma \left( \frac{\mu_*}{\sqrt{1+\frac{\pi}{8} \, \sigma^2_*}} \right) .
    \end{aligned}
\end{equation*}
It has a generalization to multi-class classification, due to \citet{gibbs1997bayesian}, i.e. for approximating 
\begin{equation} \label{eq:softmax_gauss_integral}
    p(y_* \mid x_*) = \int_{\R^C} \softmax(f_*) \, \N(f_* \mid \mu_*, \varSigma_*) \,df_* .    
\end{equation}
In this case, we approximate the resulting probability vector of length $C$ with a vector which $i$-th component is given by $\exp(\tau_i)/\sum_{j=1}^C \exp(\tau_j)$, where $\tau_j = \mu_{*j}/\sqrt{1+\pi/8 \, \varSigma_{*jj}}$ for each $j = 1, \dots, C$. This approximation ignores the correlation between logits since it only depends on the diagonal of $\varSigma_*$. Nevertheless, it yields good results even in deep learning \citep{lu2020uncertainty}, and are invaluable tools for theoretical work \citep{eschenhagen2021mola}.

\paragraph*{Laplace Bridge} The main idea of the Laplace bridge is to perform a Laplace approximation to the Dirichlet distribution by first writing it as a distribution over $\R^C$ with the help of the softmax function \citep{mackay1998choice,hennig2012kernel}. This way, Laplace approximation can be reasonably applied to approximate the Dirichlet, which can be thought as mapping the Dirichlet $\mathrm{Dir}(\alpha_*)$ to a Gaussian $\N(\mu_*, \varSigma_*)$. The pseudo-inverse of this map, mapping $(\mu_*, \varSigma_*)$ to $\alpha_*$ where for each $i = 1, \dots, C$, the $i$-th component $\alpha$ is given by the simple closed-form expression 
\begin{equation*}
    \alpha_i = \frac{1}{\varSigma_{ii}} \left( 1 - \frac{2}{C} + \frac{\exp(\mu_i)}{C^2} \sum_{j=1}^C \exp(-\mu_j) \right) ,
\end{equation*}
is the \emph{Laplace bridge}. Just like the probit approximation, the Laplace bridge ignores the correlation between logits. But, unlike all the previous approximations, it yields a \emph{full distribution} over the solutions of the softmax-Gaussian integral \eqref{eq:softmax_gauss_integral}. So, the Laplace bridge is a richer yet comparably simple approximation to the integral and is useful for many applications in deep BNNs \citep{hobbhahn2020fast}.

        \section{Further Experiments Details and Results}
        
\subsection{Laplace Comparison}
\label{app:sec:la_comparison}

\begin{figure}[t!]
   \subfloat[Hessian structure (CIFAR-10 + DA)]{\includegraphics[width=0.5\linewidth]{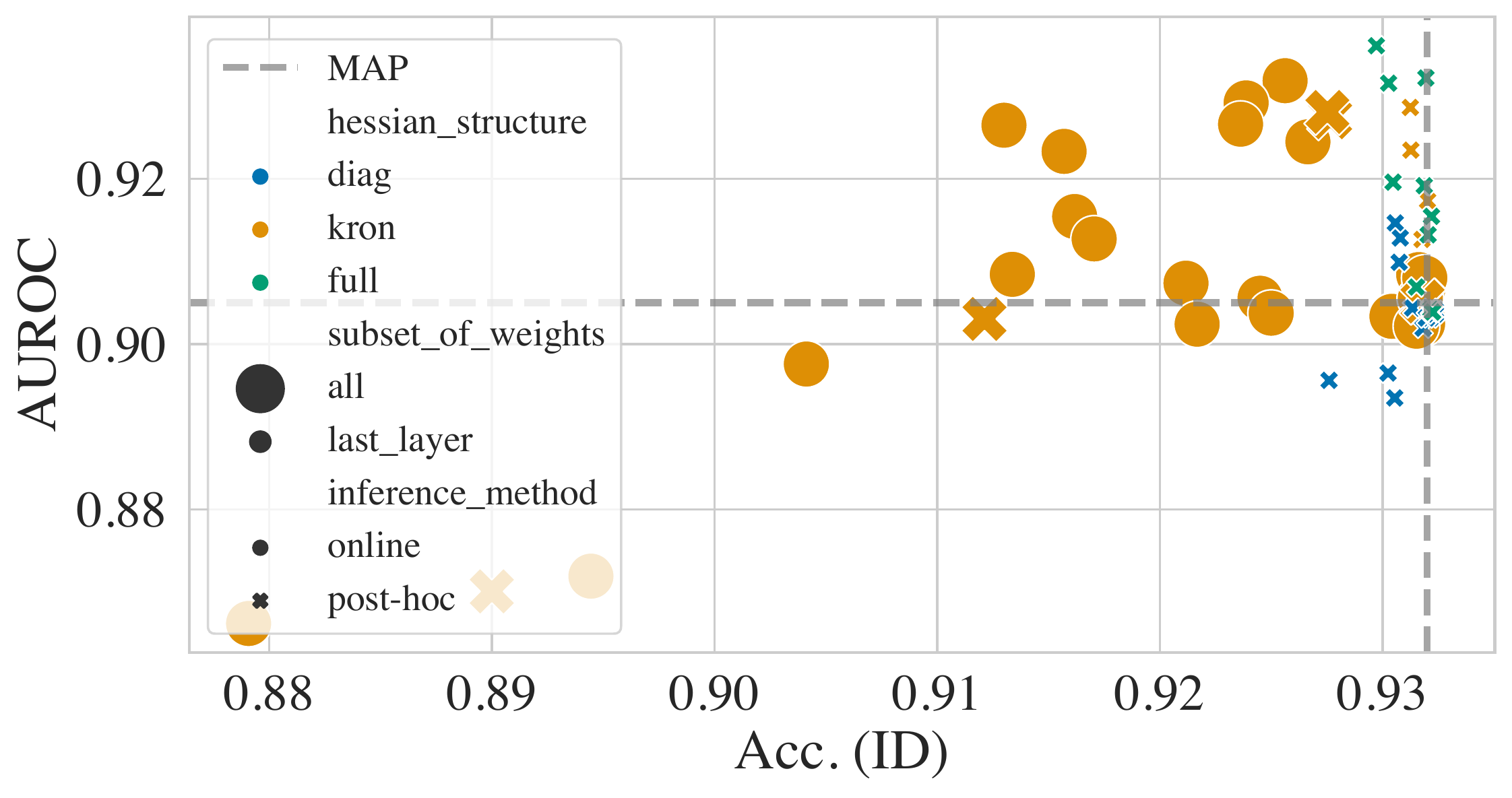}}
   \subfloat[Hessian structure (CIFAR-10)]{\includegraphics[width=0.5\linewidth]{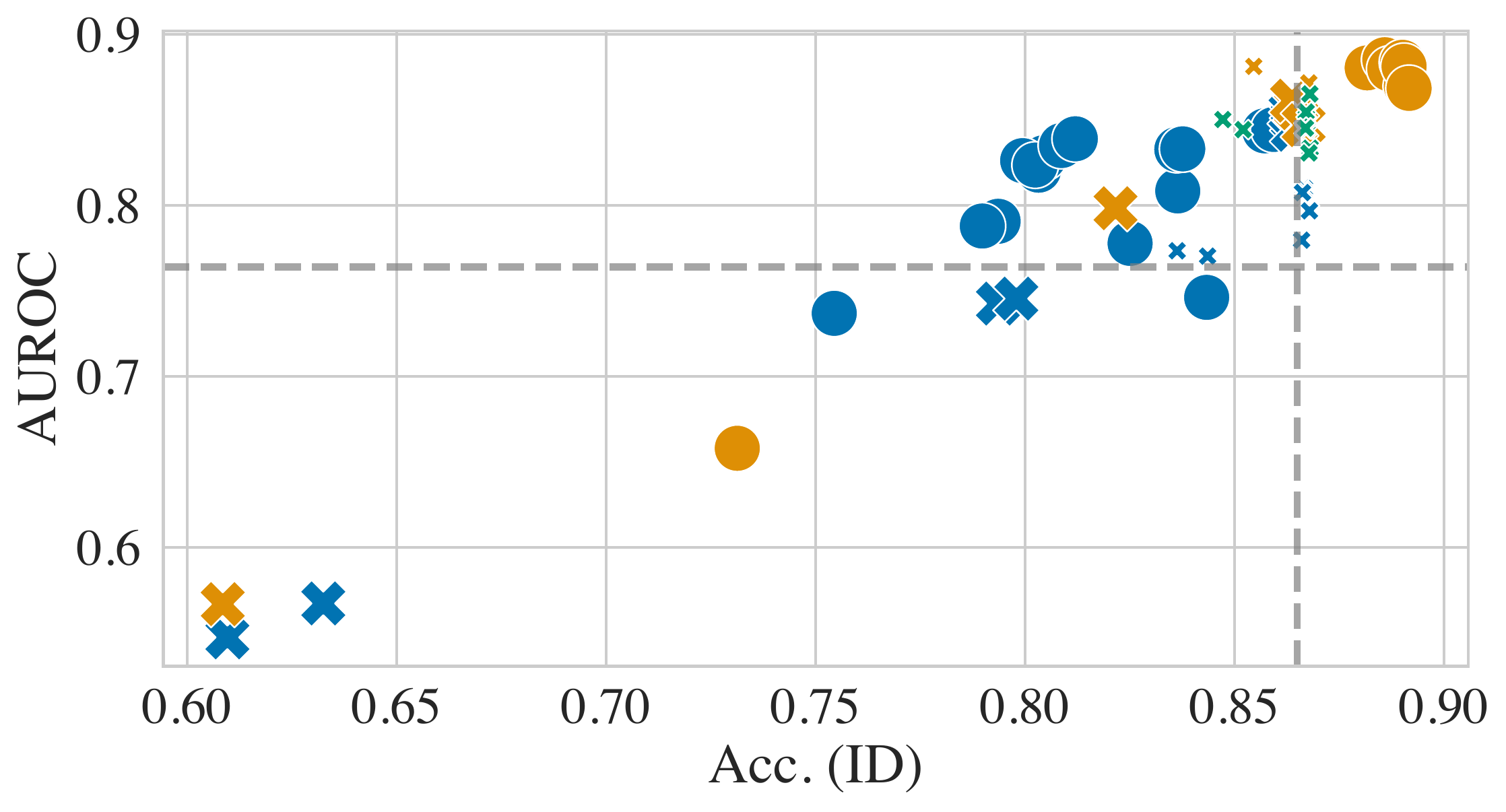}} \\
   \subfloat[Predictive approximation (CIFAR-10 + DA)]{\includegraphics[width=0.5\linewidth]{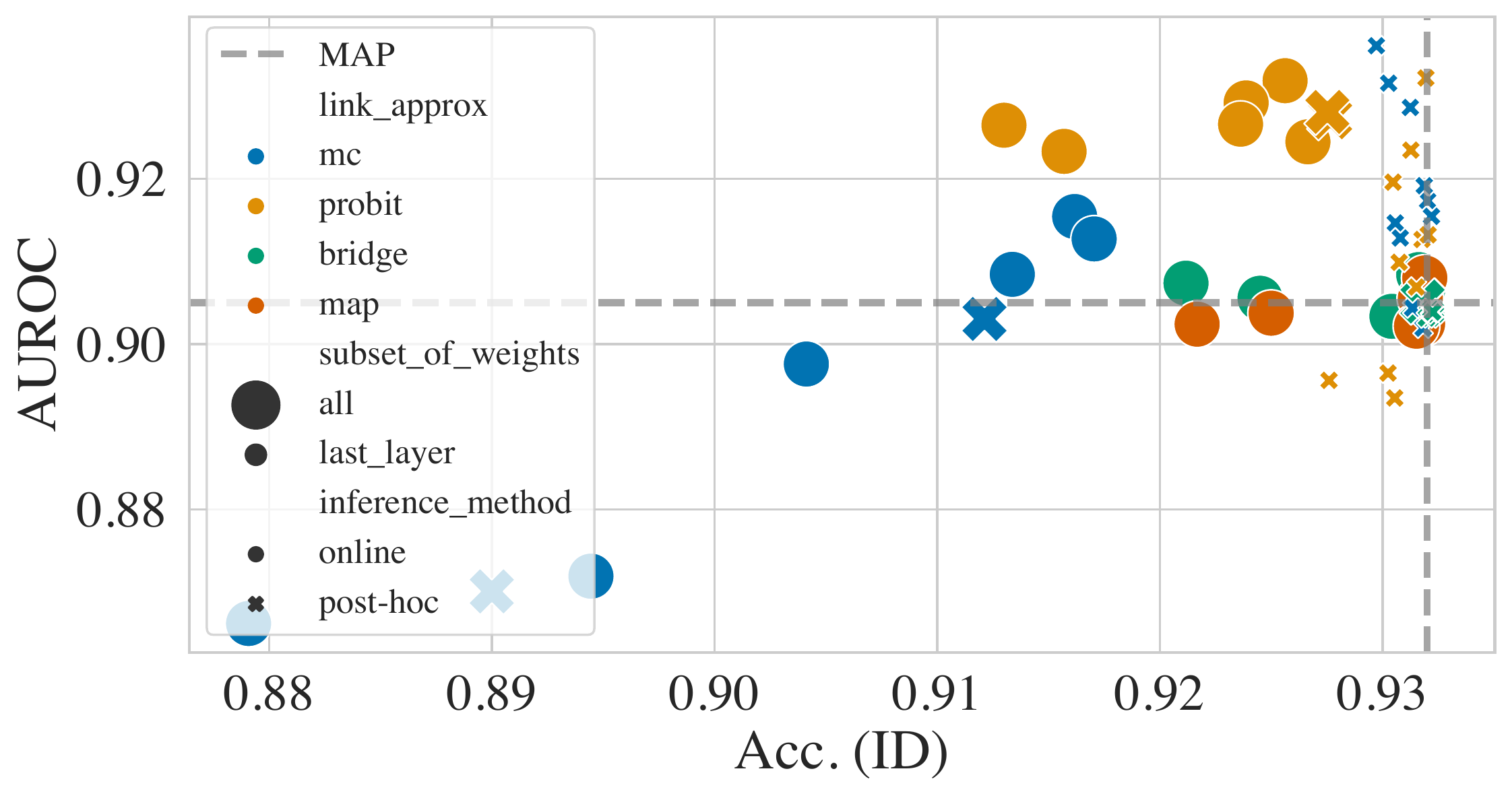}}
   \subfloat[Predictive approximation (CIFAR-10)]{\includegraphics[width=0.5\linewidth]{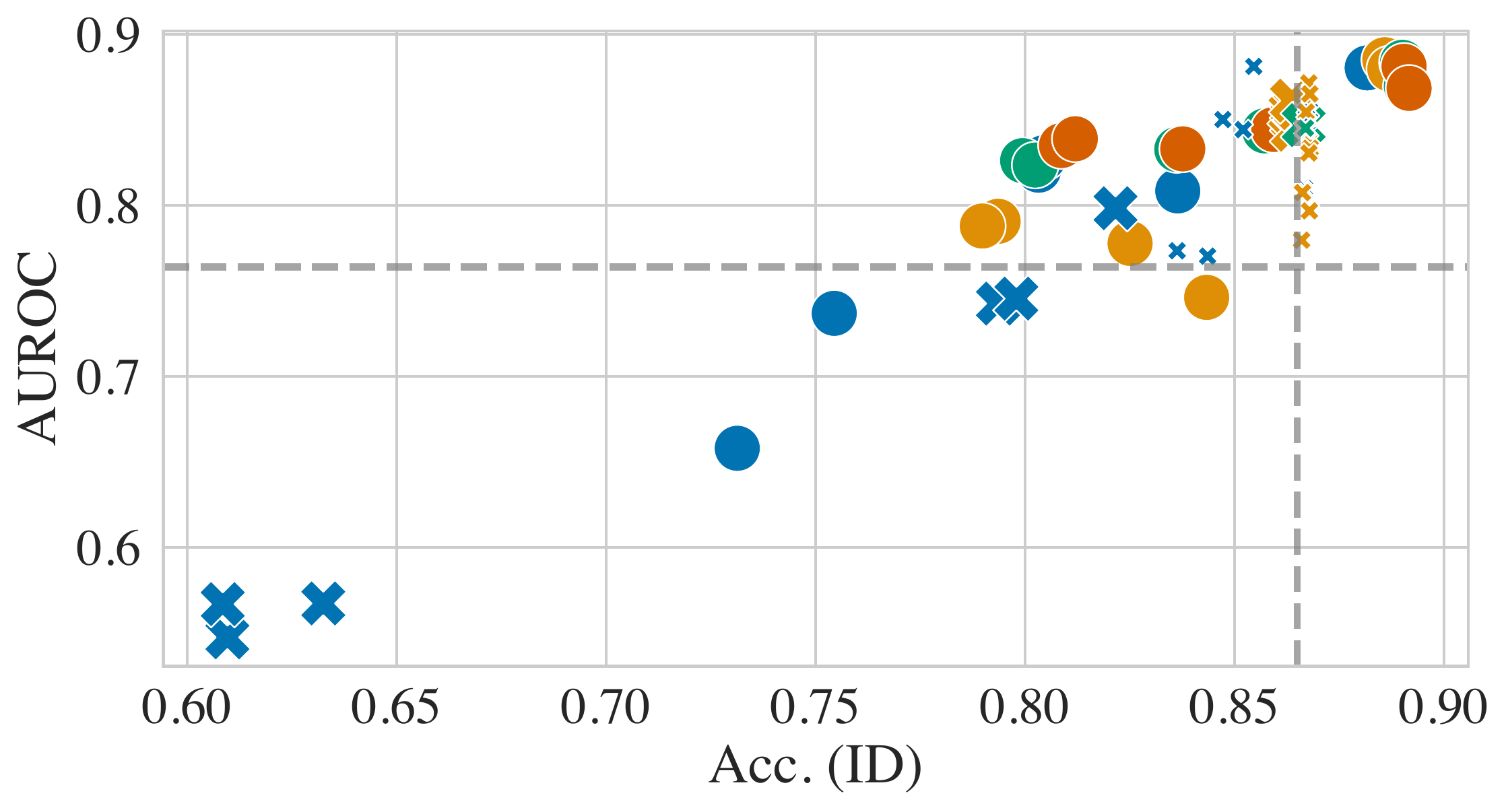}}
    \caption{Comparison of variations of the LA on the CIFAR-10 OOD experiment with ((a) and (c)) and without ((b) and (d)) data augmentation (DA).}
    \label{fig:laplace_comparison_appendix}
\end{figure}

Here, we present more detailed results of our comparison of the different variations of the Laplace approximation. We show in-distribution accuracy for CIFAR-10 using a model trained with and without data augmentation, and AUROC values averaged over the out-of-distribution datasets SVHN, LSUN, and CIFAR-100. In the first row of \Cref{fig:laplace_comparison_appendix}, we highlight the different Hessian structures with different colors; in the second row, we use color to highlight the different link approximations in the predictive distribution. We considered most combinations of the different choices for the components discussed in \Cref{subsec:ingredients}, but exclude some combinations which we have found to not work well at all, e.g. online Laplace when performing a Laplace approximation over the weights of only the last layer.
In \Cref{tab:hessian_approx_runtime}, we compare the predictive performance and runtime when using differently structured Hessian approximations.
We find that the Kronecker-factored Hessian approximations provides a good trade-off between runtime and performance.

\begin{table}[]
    \centering
    \begin{tabular}{lllll}
    \toprule
        {} & test log likelihood & test accuracy & OOD-AUROC & prediction time (s) \\
    \midrule
        \textsc{diag} & -0.302$\pm$0.005 & 0.894$\pm$0.002 & 0.832$\pm$0.011 & 29.5$\pm$0.2 \\
        \textsc{kfac} & -0.282$\pm$0.004 & 0.899$\pm$0.002 & 0.836$\pm$0.004 & 30.6$\pm$0.1 \\
        \textsc{full} & -0.285$\pm$0.004 & 0.898$\pm$0.002 & 0.876$\pm$0.003 & 62.8$\pm$1.1 \\
    \bottomrule
    \end{tabular}
    \vspace{1em}
    \caption{Qualitative comparison of different Hessian approximations. The \textsc{kfac} Hessian approximation performs similar to \textsc{full} Gauss-Newton but is almost as fast as \textsc{diag}. We use online marginal likelihood method~\citep{immer2021scalable} to train a small convolutional network on FMNIST and measure performance at test time.
    We repeat for three seeds to estimate the standard error.
    The OOD-AUROC is averaged over EMNIST, MNIST, and KMNIST. The prediction time is taken as the average over all in and out-of-distribution data sets.
    We use the MC predictive with $100$ samples.
    }
    \vspace{-1em}
    \label{tab:hessian_approx_runtime}
\end{table}

\subsection{Predictive Uncertainty Quantification}
\label{app:sec:uq_details}

\subsubsection{Training Details} 

We use LeNet \citep{lecun1998gradient} and WideResNet-16-4 \citep[WRN,][]{zagoruyko2016wide} architectures for the MNIST and CIFAR-10 experiments, respectively. We adopt the commonly-used training procedure and hyperparameter values.

\paragraph{MAP} We use Adam and Nesterov-SGD to train LeNet and WRN, respectively. The initial learning rate is $0.1$ and annealed via the cosine decay method \citep{loshchilov2016sgdr} over $100$ epochs. The weight decay is set to \num{5e-4}. Unless stated otherwise, all methods below use these training parameters.

\paragraph{DE} We train five MAP network (see above) independently to form the ensemble.

\paragraph{VB} We use the Bayesian-Torch library \citep{krishnan2020bayesiantorch} to train the network. Tha variational posterior is chosen to be the diagonal Gaussian \citep{graves2011practicalVI,blundell2015weight} and the flipout estimator \citep{wen2018flipout} is employed. The prior precision is set to \num{5e-4} to match the MAP-trained network, while the KL-term downscaling factor is set to $0.1$, following \citep{osawa2019practical}.

\paragraph{CSGHMC} We use the publicly available code provided by the original authors \citep{zhang2020csgmcmc}.\footnote{\url{https://github.com/ruqizhang/csgmcmc}} We use their default (i.e. recommended) hyperparameters.

\paragraph{SWAG} For the SWAG baseline, we follow \citet{maddox2019simple} and run stochastic gradient descent with a constant learning rate on the pre-trained models to collect one model snapshot per epoch, for a total of 40 snapshots. At test time, we then make predictions by using 30 Monte Carlo samples from the posterior distribution; we correct the batch normalization statistics of each sample as described in \citet{maddox2019simple}.
To tune the constant learning rate, we used the same approach as in \citet{eschenhagen2021mola}, combining a grid search with a threshold on the mean confidence.
For MNIST, we defined the grid to be the set $\{$ 1e-1, 5e-2, 1e-2, 5e-3, 1e-3 $\}$, yielding an optimal value of 1e-2.
For CIFAR-10, searching over the same grid suggested that the optimal value lies between 5e-3 and 1e-3; another, finer-grained grid search over the set $\{$ 5e-3, 4e-3, 3e-3, 2e-3, 1e-3 $\}$ then revealed the best value to be 2e-3.

\paragraph{Other baselines} Our choice of baselines is based on the most common and best performing methods of recent Bayesian DL papers.
Despite its popularity, \textbf{Monte Carlo (MC) dropout} \citep{gal2016dropout} has been shown to underperform compared to more recent methods (see e.g. \citet{ovadia2019can}). A recent VI method called \textbf{Variational Online Gauss-Newton (VOGN)} \citep{osawa2019practical} also seems to underperform. For example, Fig.~5 of \citet{osawa2019practical} shows that on OOD detection with CIFAR-10 vs. SVHN, MC-dropout and VOGN only achieve AUROC$\uparrow$ values of $81.9$ and $80.0$, respectively, while last-layer-LA obtains a substantially better value of $91.9$ (they use ResNet-18, which is comparable to our model).

\subsubsection{Detailed Results}
\begin{figure}[t!]
    \subfloat[MNIST-R Brier $\downarrow$]{\includegraphics[width=0.25\linewidth]{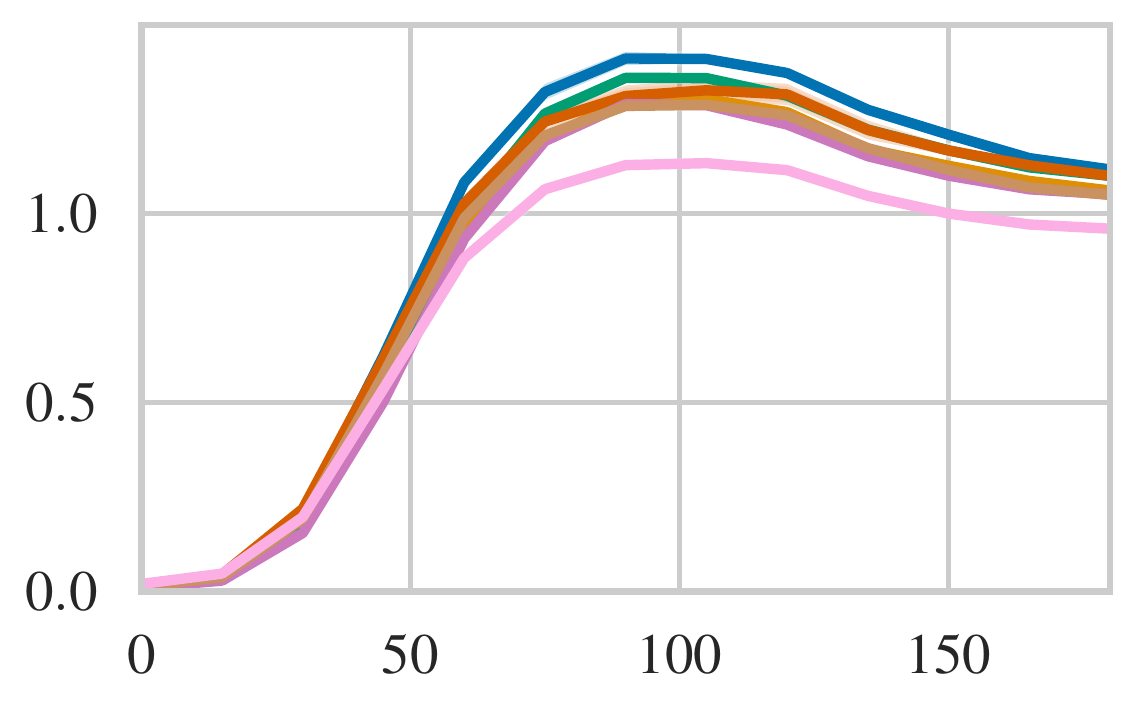}}
    \subfloat[MNIST-R Acc. $\uparrow$]{\includegraphics[width=0.25\linewidth]{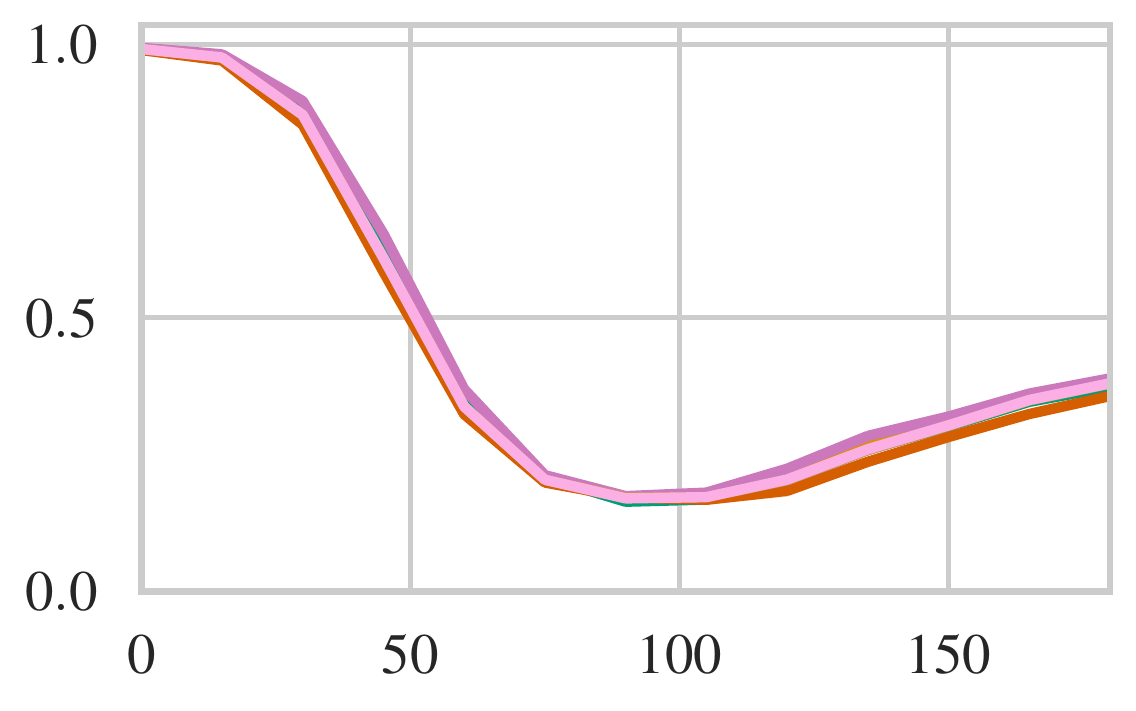}}
    \subfloat[CIFAR10-C Brier $\downarrow$]{\includegraphics[width=0.25\linewidth]{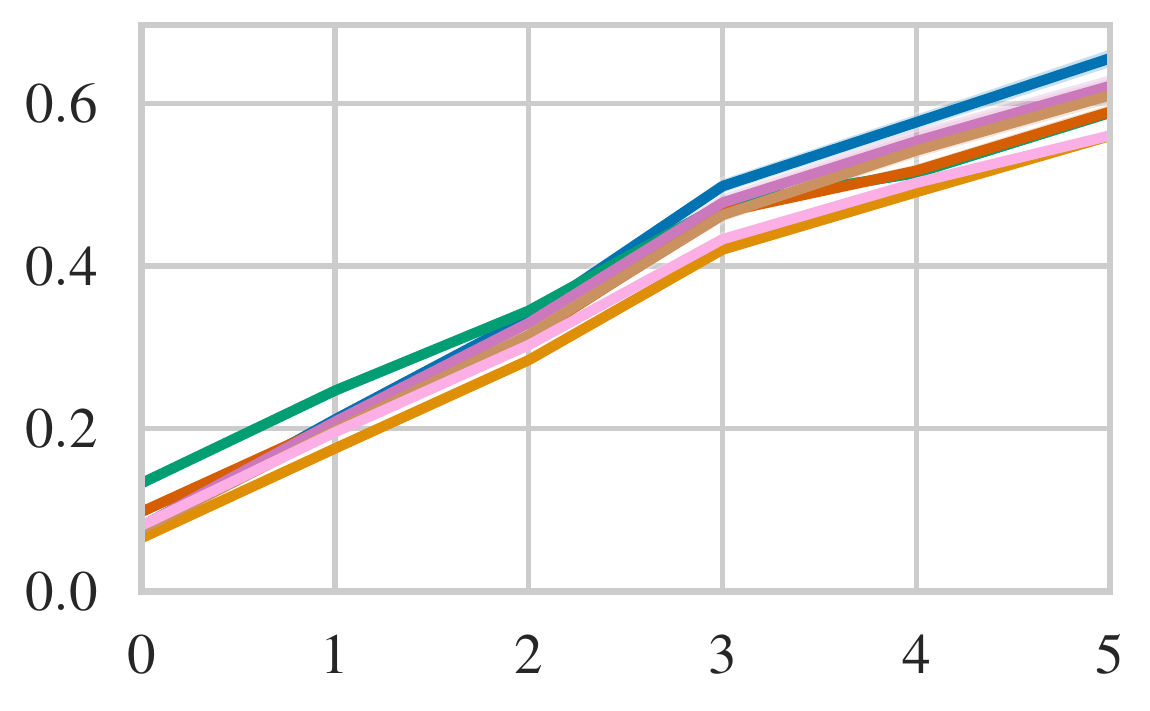}}
    \subfloat[CIFAR10-C Acc. $\uparrow$]{\includegraphics[width=0.25\linewidth]{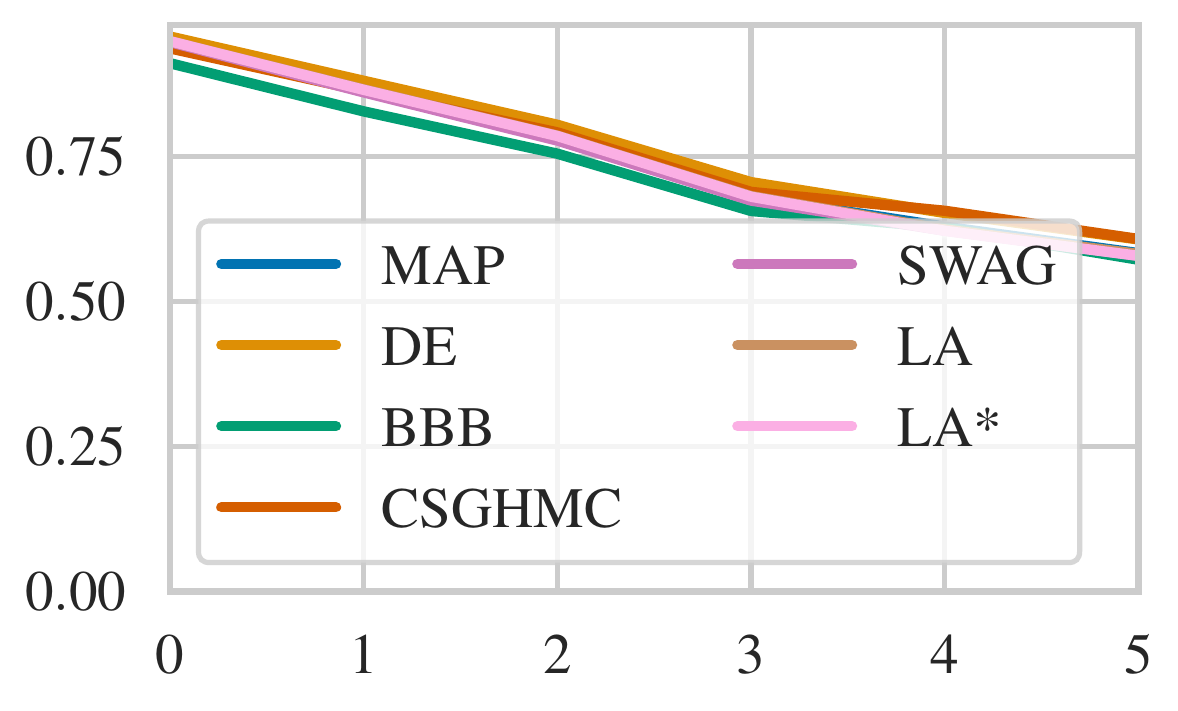}}

    \caption{Dataset shift on the Rotated-MNIST (top) and Corrupted-CIFAR-10 datasets (bottom).}
    \label{fig:dataset_shift_appendix}
\end{figure}

We show the Brier score and accuracy as a function of shift intensity in \cref{fig:dataset_shift_appendix}. Moreover, we provide the detailed (i.e. non-averaged) OOD detection results in \cref{tab:ood_mnist_appendix,tab:ood_cifar10_appendix}.

\begin{table}[t]
    \caption{MNIST OOD detection results.}
    \label{tab:ood_mnist_appendix}

    \centering
    \small

    \begin{tabular}{lrrrrrrrr}
        \toprule

        & & \multicolumn{3}{c}{\bf Confidence $\downarrow$} & & \multicolumn{3}{c}{\bf AUROC $\uparrow$} \\

        \cmidrule(r){3-5} \cmidrule(r){7-9}

        \textbf{Methods} & & {\bf EMNIST} & {\bf FMNIST} & {\bf KMNIST} & & {\bf EMNIST} & {\bf FMNIST} & {\bf KMNIST} \\

        \midrule

        MAP & & 83.6$\pm$0.3 & 64.2$\pm$0.5 & 77.3$\pm$0.3 & & 93.5$\pm$0.3 & 98.9$\pm$0.0 & 97.0$\pm$0.1 \\
        DE & & 75.8$\pm$0.2 & 55.4$\pm$0.4 & 65.9$\pm$0.3 & & 95.1$\pm$0.0 & 99.2$\pm$0.0 & 98.3$\pm$0.0 \\
        BBB & & 79.1$\pm$0.4 & 67.5$\pm$1.6 & 73.1$\pm$0.4 & & 92.3$\pm$0.2 & 98.2$\pm$0.2 & 97.0$\pm$0.2 \\
        CSGHMC & & 76.2$\pm$1.6 & 63.6$\pm$1.9 & 67.9$\pm$1.5 & & 93.4$\pm$0.2 & 97.7$\pm$0.2 & 97.1$\pm$0.1 \\
        SWAG & & 64.9$\pm$0.3 & 84.0$\pm$0.2 & 78.5$\pm$0.3 & & 98.9$\pm$0.0 & 93.6$\pm$0.3 & 97.1$\pm$0.1 \\
        LA & & 74.8$\pm$0.4 & 58.8$\pm$0.5 & 69.0$\pm$0.4 & & 93.4$\pm$0.3 & 98.5$\pm$0.1 & 96.6$\pm$0.1 \\
        LA* & & 62.0$\pm$0.5 & 49.6$\pm$0.6 & 56.7$\pm$0.5 & & 94.3$\pm$0.2 & 98.3$\pm$0.1 & 96.6$\pm$0.2 \\

        \bottomrule
    \end{tabular}
\end{table}

\begin{table}[t]
    \caption{CIFAR-10 OOD detection results.}
    \label{tab:ood_cifar10_appendix}

    \centering
    \small

    \begin{tabular}{lrrrrrrrr}
        \toprule

        & & \multicolumn{3}{c}{\bf Confidence $\downarrow$} & & \multicolumn{3}{c}{\bf AUROC $\uparrow$} \\

        \cmidrule(r){3-5} \cmidrule(r){7-9}

        \textbf{Methods} & & {\bf SVHN} & {\bf LSUN} & {\bf CIFAR-100} & & {\bf SVHN} & {\bf LSUN} & {\bf CIFAR-100} \\

        \midrule

        MAP & & 77.5$\pm$2.9 & 71.3$\pm$0.6 & 79.3$\pm$0.1 & & 91.8$\pm$1.2 & 94.5$\pm$0.2 & 90.1$\pm$0.1 \\
        DE & & 62.8$\pm$0.7 & 62.6$\pm$0.4 & 70.8$\pm$0.0 & & 95.4$\pm$0.2 & 95.3$\pm$0.1 & 91.4$\pm$0.1 \\
        BBB & & 60.2$\pm$0.7 & 53.8$\pm$1.1 & 63.8$\pm$0.2 & & 88.5$\pm$0.4 & 91.9$\pm$0.4 & 84.9$\pm$0.1 \\
        CSGHMC & & 69.8$\pm$0.8 & 65.2$\pm$0.8 & 73.1$\pm$0.1 & & 91.2$\pm$0.3 & 92.6$\pm$0.3 & 87.9$\pm$0.1 \\
        SWAG & & 69.3$\pm$4.0 & 62.2$\pm$2.3 & 73.0$\pm$0.4 & & 91.6$\pm$1.3 & 94.0$\pm$0.7 & 88.2$\pm$0.5 \\
        LA & & 70.6$\pm$3.2 & 63.8$\pm$0.5 & 72.6$\pm$0.1 & & 92.0$\pm$1.2 & 94.6$\pm$0.2 & 90.1$\pm$0.1 \\
        LA* & & 58.0$\pm$3.1 & 50.0$\pm$0.5 & 59.0$\pm$0.1 & & 91.9$\pm$1.3 & 95.0$\pm$0.2 & 90.2$\pm$0.1 \\

        \bottomrule
    \end{tabular}
\end{table}

\subsubsection{Additional Details on Wall-clock Time Comparison}
Concerning the wall-clock time comparison in \cref{fig:costs}, we would like to clarify that for LA, we consider the default configuration of \libname{}. As the default LA variant uses the closed-form probit approximation to the predictive distribution and therefore neither requires Monte Carlo (MC) sampling nor multiple forward passes, the wall-clock time for making predictions is essentially the same as for MAP.
This is contrast to the baseline methods, which are significantly more expensive at prediction time due to the need for MC sampling (VB, SWAG) or forward passes through multiple model snapshots (DE, CSGHMC).

Importantly, note that is an advantage exclusive to our implementation of LA (i.e. with a GGN/Fisher Hessian approximation or with the last-layer LA) that it can be used without sampling (i.e.~using the probit or Laplace bridge predictive approximations).
This kind of approximation is incompatible with the other baselines (i.e.~DE, CSGHMC, SWAG, and VB) since these methods just yield samples/distributions over weights while our LA variants implicitly yield a Gaussian distribution over logits due to the linearization of the NN induced by the use of the GGN/Fisher (see \citet{immer2020improving} for details) or the use of only the last layer. While one could still apply linearization to other methods, this would not be theoretically justified, in contrast to GGN-/last-layer-LA. 

Finally, the reason we benchmark our deterministic, probit-based version is that we found it to consistently perform on par or better than MC sampling. If we predict with the LA using MC samples on the logits, the runtime is only around 20\% slower than the deterministic probit approximation, which is still significantly faster than all other methods.

In summary, we believe that the ability to obtain calibrated predictions with a single forward-pass is a critical and distinctive advantage of the LA over almost all other Bayesian deep learning and ensemble methods.

\subsection{\texttt{WILDS} Experiments}
\label{sec:wilds_details}

For this set of experiments, we use \texttt{WILDS} \cite{koh2020wilds}, a recently proposed benchmark of realistic distribution shifts encompassing a variety of real-world datasets across different data modalities and application domains.
In particular, we consider the following \texttt{WILDS} datasets: 
\begin{itemize}
    \item \texttt{Camelyon17}: Tumor classification (binary) of histopathological tissue images across different hospitals (ID vs.\ OOD) using a DenseNet-121 model (10 seeds).
    \item \texttt{FMoW}: Building / land use classification (62 classes) of satellite images across different times and regions (ID vs.\ OOD) using a DenseNet-121 model (3 seeds).
    \item \texttt{CivilCommments}: Toxicity classification (binary) of online text comments across different demographic identities (ID vs.\ OOD) using a DistilBERT-base-uncased model (5 seeds).
    \item \texttt{Amazon}: Sentiment classification (5 classes) of product reviews across different reviewers (ID vs.\ OOD) using a DistilBERT-base-uncased model (3 seeds).
    \item \texttt{PovertyMap}: Asset wealth index regression (real-valued) across different countries and rural/urban areas (ID vs.\ OOD) using a ResNet-18 model (5 seeds).
\end{itemize}

\begin{figure}[t!]
    \centering
    \includegraphics[width=\textwidth]{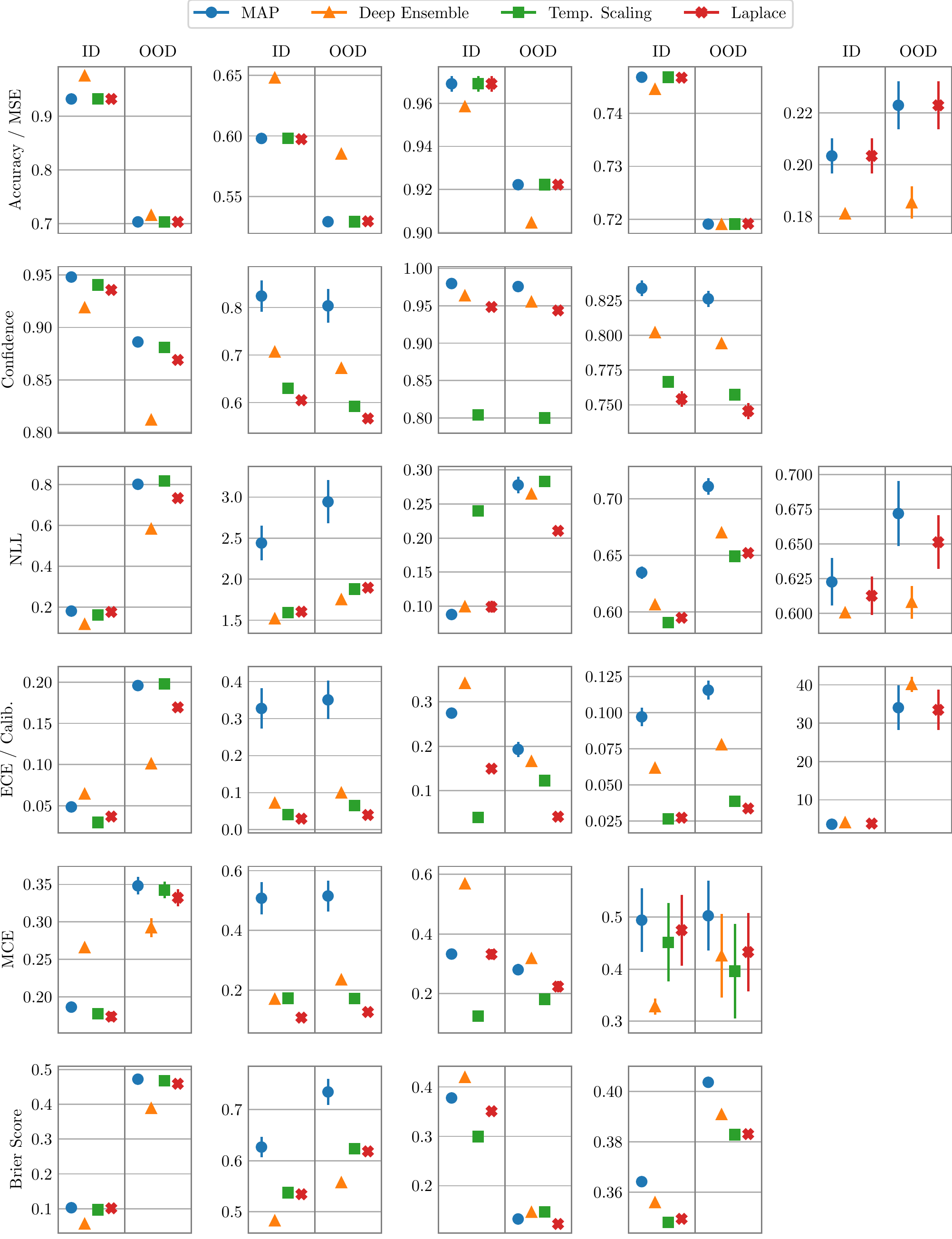}
    
    \hspace{.03\linewidth}
    \subfloat[\label{subfig:camelyon17_full} \texttt{Camelyon17}]{\hspace{.19\linewidth}}
    \subfloat[\label{subfig:fmow_full} \texttt{FMoW}]{\hspace{.19\linewidth}}
    \subfloat[\label{subfig:civilcomments_full} \texttt{CivilComments}]{\hspace{.20\linewidth}}
    \subfloat[\label{subfig:amazon_full} \texttt{Amazon}]{\hspace{.19\linewidth}}
    \subfloat[\label{subfig:poverty_full} \texttt{PovertyMap}]{\hspace{.19\linewidth}}

    \caption{Assessing real-world distribution shift robustness on five datasets from the \texttt{WILDS} benchmark \cite{koh2020wilds}, covering different data modalities, model architectures, and output types; see text for details.
    We report means $\pm$ standard errors of several metrics (from top to bottom): accuracy (for classification) or mean squared error (for regression), confidence (only for classification), negative log-likelihood, ECE (for classification) or regression calibration error \cite{kuleshov2018accurate}, mean calibration error (only for classification), and Brier score (only for classification).
    The in-distribution (left panels) and OOD (right panels) dataset splits correspond to different domains (e.g.\ hospitals for \texttt{Camelyon17}).}
    \label{fig:wilds_full}
\end{figure}

Please refer to the original paper for more details on this benchmark and the above-mentioned datasets.
All reported results in \cref{fig:wilds} and \cref{fig:wilds_full} show the mean and standard error across as many seeds as there are provided with the original paper (see the list of datasets above for the exact numbers).

For the last-layer Laplace method, we use either a KFAC or full covariance matrix (depending on the size of the last layer; in particular, we use a KFAC covariance for \texttt{FMoW} and full covariances for all other datasets) and the linearized Monte Carlo predictive distribution with 10,000 samples.

For the deep ensemble, we simply the aggregate the pre-trained models provided by the original paper\footnote{See \url{https://worksheets.codalab.org/worksheets/0x52cea64d1d3f4fa89de326b4e31aa50a} for the complete list of models.}
This yields ensembles of 5 neural network models, which is a commly-used ensemble size \citep{ovadia2019can}.
Since these models were trained in different ways (e.g.\ using different domain generalization methods, see \cite{koh2020wilds} for details), their combinations can be viewed as \emph{hyperparameter ensembles} \cite{wenzel2020hyperparameter}.

Note that the temperature scaling baseline is only applicable for classification tasks, and therefore we do not report it for the \texttt{PovertyMap} regression dataset.

We tune the temperature parameter for temperature scaling, the prior precision parameter for Laplace, and the noise standard deviation parameter for regression (i.e.\ for the \texttt{PovertyMap} dataset) by minimizing the negative log-likelihood on the in-distribution validation sets provided with WILDS.

Finally, \cref{fig:wilds_full} shows an extended version of the results reported in \cref{fig:wilds}, which additionally reports the following metrics: accuracy (for classification) or mean squared error (for regression), confidence (only for classification), mean calibration error (only for classification), and Brier score (only for classification).
The overall conclusion here is the same as for \cref{fig:wilds}, namely that Laplace is significantly better calibrated than MAP, and competitive with temperature scaling and ensembles, especially on the OOD splits.
Note that the differences in accuracies of the ensemble stem from the different training procedures of the ensemble members (which sometimes achieve higher and sometimes lower accuracy), as mentioned above.

\subsection{Further Details on the Continual Learning Experiment}
\label{app:sec:cl_details}

We benchmark Laplace approximations in the Bayesian continual learning setting on the \emph{permuted MNIST} benchmark which consists of $10$ consecutive tasks where each task is a permutation of the pixels of the MNIST images.
Following common practice~\citep{ritter2018online, nguyen2018variational, osawa2019practical}, we use a $2$-hidden layer MLP with $100$ hidden units each and $28\times 28= 784$ input dimensions and $10$ output dimensions for the MNIST classes.
We adopt the implementation of the continual learning task and the model by \citet{pan2020continual}.\footnote{The code is avilable at \url{https://github.com/team-approx-bayes/fromp}.}
In the following, we will briefly outline the Bayesian approach to continual learning~\citep{nguyen2018variational} and explain how a diagonal and KFAC Laplace approximation can be employed in this setting.
Further, we describe how this can be combined with the evidence framework to update the prior online alleviating the need for a validation set, which is unlikely to be available in real continual learning scenarios.

\subsubsection{Bayesian Approach to Continual Learning}
The Bayesian approach to continual learning can be simply described as iteratively updating the posterior after each task.
We are given $T$ data sets $\D := \{\D_t \}_{t=1}^T$ and have a neural network with parameters $\theta$.
In line with the standard supervised learning setting outlined in \cref{sec:laplace},
we have a prior on parameters $p(\theta)=\mathcal{N}(\theta; 0, \gamma^2 I)$ and a likelihood $p(\D \mid \theta)$ realized by a neural network.
The posterior on the parameters after all tasks is then
\begin{equation}
    p(\theta \mid \D) \propto p(\D_T \mid \theta) \times \ldots \times \underbrace{p(\D_2 \mid \theta) \times \underbrace{p(\D_1 \mid \theta) \times p(\theta)}_{\propto p(\theta \mid \D_1)}}_{\propto p(\theta \mid \D_1, \D_2)}.
    \label{eq:continual_factorization}
\end{equation}
This factorization gives rise to a recursion to update the posterior after $t-1$ data sets to the posterior after $t$ data sets:
\begin{equation}
    p(\theta \mid \D_1, \ldots, \D_{t}) \propto p(\D_t \mid \theta) p(\theta \mid \D_1, \ldots, \D_{t-1}).
    \label{eq:continual_recursion}
\end{equation}
The normalizer for each update in \cref{eq:continual_recursion} is given by the marginal likelihood $p(\D_t \mid \D_1, \ldots, \D_{t-1})$ and we will use it for optimizing the variance $\gamma^2$ of $p(\theta)$.
Incorporating a new task is the same as Bayesian inference in the supervised case but with an updated prior, i.e., the prior is the previous posterior distribution on $\theta$.
The Laplace approximation provides one way to approximately infer the posterior distributions after each task~\citep{huszar2017quadratic, ritter2018online, pan2020continual}.
Alternatively, variational inference can be used~\citep{nguyen2018variational, osawa2019practical}.

\subsubsection{The Laplace Approximation for Continual Learning}
The Laplace approximation facilitates the recursive updates~(\cref{eq:continual_recursion}) that arise in continual learning.
In this context, it was first suggested with a diagonal Hessian approximation by~\citet[EWC]{kirkpatrick2017overcoming} and \citet{huszar2017quadratic} corrected their updates.
\citet{ritter2018online} greatly improved the performance by using a KFAC Hessian approximation instead of a diagonal.
The Laplace approximation to the posterior after observing task $t$ is a Gaussian $\N(\theta_\map^{(t)}, \Sigma^{(t)})$
We obtain $\theta_\map$ by optimizing the unnormalized log posterior distribution on $\theta$ as annotated in \cref{eq:continual_factorization} for every task, one after another.
The Hessian of the same unnormalized log posterior also specifies the posterior covariance $\Sigma^{(t)}$:
\begin{equation}
    \Sigma^{(t)} = \Big( \underbrace{\nabla_\theta^2 \log p(\D_t \mid \theta)\vert_{\theta_\map^{(t)}}}_{\text{log likelihood Hessian}}\, +\, \underbrace{\textstyle\sum_{t'=1}^{t-1} \nabla_\theta^2 \log p(\D_{t'} \mid \theta)\vert_{\theta_\map^{(t')}} }_{\text{previous log likelihood Hessians}} \, + \, \underbrace{\gamma^{-2} I}_{\text{log prior Hessian}} \Big)^{-1}.
    \label{eq:continual_posterior_cov}
\end{equation}
This summation over Hessians is typically intractable for neural networks with large parameter vectors $\theta$ and hence diagonal or KFAC approximations are used~\citep{kirkpatrick2017overcoming, huszar2017quadratic, ritter2018online}.
For the diagonal version, the addition of Hessians and log prior is exact.
For the KFAC version, we follow the alternative suggestion by \citet{ritter2018online} and add up Kronecker factors which is an approximation to the sum of Kronecker products.
However, this approximation is what underlies KFAC even in the supervised learning case where we add up factors per data point over the entire data set.
Lastly, we adapt $\gamma$ during training on each task $t$ by optimizing the marginal likelihood $p(\D_t \mid \D_1, \ldots \D_{t-1})$, i.e., by differentiating it with respect to $\gamma$.
This can be done by computing the eigendecomposition of the summed Kronecker factors~\citep{immer2021scalable} and allows us to 1) adjust the regularization suitably per task and 2) avoid setting a hyperparameter thereby alleviating the need for validation data.

\subsection{Comparison of Memory Complexity}
\label{app:sec:memory}
\cref{tab:memory} compares the theoretical memory complexity and actual memory footprint (of a Wide ResNet 16-4 on CIFAR-10) of the different methods.
\begin{table}[t]
    \caption{The memory complexities of all methods in $\mathcal{O}$ notation.
    To get a better idea of what these complexities translate to in practice, we also report the actual memory footprints (in megabytes) of a Wide ResNet 16-4 (WRN) on CIFAR-10.
    Here, $M$ denotes the number of model parameters, $H$ denotes the number of neurons in the last layer, $K$ denotes the number of model outputs, $R$ denotes the number of SWAG snapshots, $S$ denotes the number of CSGHMC samples, and $N$ denotes the number of deep ensemble (DE) members. Mean-field variational inference (VB) has a complexity of $2M$ as it needs to store a variance vector of size $M$ in addition to the mean vector of size $M$. For the actual memory footprints, we assume $R = 40$ SWAG snapshots, $S = 12$ CSGHMC samples, and $N = 5$ ensemble members, which are the hyperparameters recommended in the original papers (and therefore also used in our experiments).
    It can be seen that the proposed default KFAC-last-layer approximation poses a small memory overhead of $\mathcal{O}(H^2 + K^2)$ on top of the MAP estimate. }
    \label{tab:memory}
    \vspace{0.5em}

    \centering
    \fontsize{8}{9}\selectfont
    \renewcommand{\tabcolsep}{4pt}
    \begin{sc}
    \begin{tabular}{lccrccccrcr}
        \toprule
            Method & Mem. Complexity & WRN on CIFAR-10 \\
            \midrule
            MAP & $M$ & 11MB \\
            \textbf{LA} & $\mathbf{M + H^2 + K^2}$ & \textbf{12MB} \\
            VB & $2M$ & 22MB \\
            DE & $NM$ & 55MB \\
            CSGHMC & $SM$ & 132MB \\
            SWAG & $RM$ & 440MB \\
        \bottomrule
    \end{tabular}
    \end{sc}
\end{table}
    \end{appendices}

\end{document}